\documentclass{article} 
\usepackage{iclr2026_conference,times}
\usepackage{graphicx}

\usepackage{amsmath,amsfonts,bm}

\def\eqref#1{equation~\ref{#1}}

\def\1{\bm{1}}

\DeclareMathAlphabet{\mathsfit}{\encodingdefault}{\sfdefault}{m}{sl}
\SetMathAlphabet{\mathsfit}{bold}{\encodingdefault}{\sfdefault}{bx}{n}

\usepackage{hyperref}
\usepackage{url}
\usepackage{amsmath}
\usepackage{algorithm}
\usepackage{algpseudocode}
\usepackage{multirow}
\usepackage{graphicx}
\usepackage{booktabs}
\usepackage{enumitem}
\usepackage{xcolor} 
\usepackage[dvipsnames]{xcolor}  
\usepackage{colortbl} 
\usepackage{amssymb} 
\definecolor{top1}{HTML}{F7CDC8}
\definecolor{top2}{HTML}{D1EDF9}
\definecolor{top3}{HTML}{FDF1D0}

\hypersetup{
    colorlinks=true, 
    citecolor=blue,  
}

\title{WorldEdit: Towards Open-World Image Editing with a Knowledge-Informed Benchmark}

\iclrfinalcopy

\author{
Wang Lin\textsuperscript{\rm 1}\quad
Feng Wang\textsuperscript{\rm 2}\quad
Majun Zhang\textsuperscript{\rm 1}\quad
Wentao Hu\textsuperscript{\rm 3}\quad
Tao Jin\textsuperscript{\rm 1}\\
\,\textbf{Zhou Zhao}\textsuperscript{\rm 1}\quad
\textbf{Fei Wu}\textsuperscript{\rm 1}\quad
\textbf{Jingyuan Chen}\textsuperscript{\rm 1}\thanks{ Corresponding author.}\quad
\textbf{Alan Yuille}\textsuperscript{\rm 2}\quad
\textbf{Sucheng Ren}\textsuperscript{\rm 2}\\
\textsuperscript{\rm 1}Zhejiang University \quad 
\textsuperscript{\rm 2}Johns Hopkins University\\
\textsuperscript{\rm 3}Nanyang Technological University
\\
\tt\small linwanglw@zju.edu.cn\quad
}

\usepackage{xspace}

\newcommand{\eg}{\textit{e.g.}\@\xspace}

\begin{document}

\maketitle

\begin{abstract}

Recent advances in image editing models have demonstrated remarkable capabilities in executing explicit instructions, such as attribute manipulation, style transfer, and pose synthesis.    However, these models often face challenges when dealing with implicit editing instructions, which describe the cause of a visual change without explicitly detailing the resulting outcome.    These limitations arise because existing models rely on uniform editing strategies that are not equipped to handle the complex world knowledge and reasoning required for implicit instructions.    
To address this gap, we introduce \textbf{WorldEdit}, a dataset specifically designed to enable world-driven image editing.    
WorldEdit consists of high-quality editing samples, guided by paraphrased instructions that align with real-world causal logic. 
Furthermore, we provide \textbf{WorldEdit-Test} for evaluating the existing model's performance on causal editing scenarios.
With WorldEdit, we use a two-stage training framework for fine-tuning models like Bagel, integrating with a causal verification reward.   
Our results show that the proposed dataset and methods significantly narrow the gap with GPT-4o and Nano-Banana, demonstrating competitive performance not only in instruction following but also in knowledge 
plausibility, where many open-source systems typically struggle. See \href{https://worldedit0.github.io/WorldEdit/}{Project Page}. 
\end{abstract}

\section{Introduction}

\begin{figure*}
  \centering
  \includegraphics[scale=0.39]{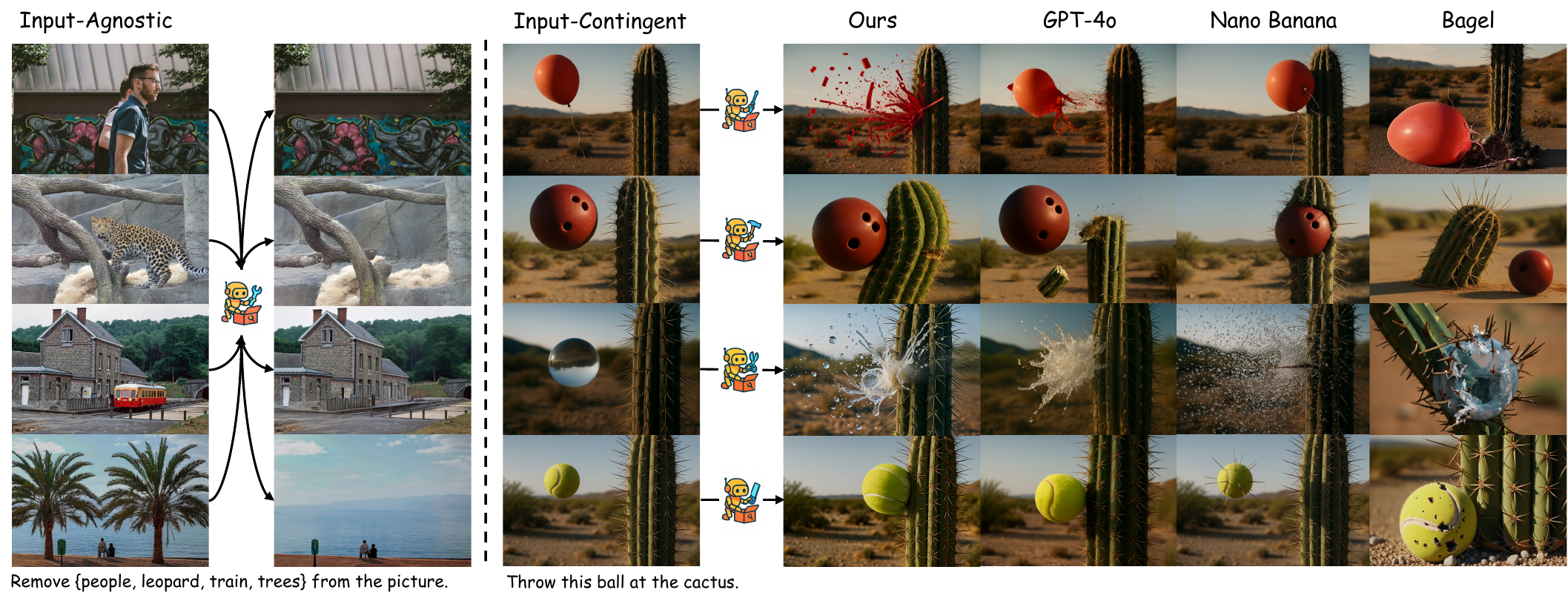}
\vspace{-1.25em}
\caption{Unlike traditional image editing (\textit{left}), which adopts a uniform editing strategy for different editing objects, world editing (\textit{right}) needs to take into account the nature of the editing objects in the real world and produce editing results that conform to causal logic.}
\label{intro}
\vspace{-1.25em}
\end{figure*}

In recent years, image editing models~\citep{simsar2024uip2p,zhu2025kv,simsar2025lime} have made remarkable progress, demonstrating excellent performance on tasks with explicit instructions—such as attribute modification~\citep{cao2023masactrl,xu2023inversion}, style transfer~\citep{chung2024style,wang2023stylediffusion}, and pose synthesis~\citep{yin2025grpose,shen2023advancing}. However, as illustrated in Figure~\ref{intro}, when confronted with implicit editing instructions, which only provide the cause of a visual change without explicitly describing the resulting visual outcome, most existing models still exhibit significant limitations in editing quality.

An intuitive workaround is to paraphrase implicit instructions into more explicit editing prompts. Yet, as shown in Figure~\ref{fig:rewrite}, we observe that even when using paraphrasing to convey editing intent to pre-trained generative models, the editing results of most models are still quite poor. On the one hand, the visual outcomes implied by such instructions are often highly complex and require accurate world knowledge to realize. For instance, the instruction ``a water balloon hits a cactus'' entails visual effects (\eg, splashing trajectories of water droplets) that must adhere to physical laws and object interaction logic. How to generate high-quality textual prompts based on such knowledge to describe the visual outcome in detail remains an open problem~\citep{deng2025emerging}.

On the other hand, even giving paraphrased instructions, many visual expressions remain challenging for pre-trained generative models to follow and render, like the single-sided structure of a Möbius strip or scattering pattern of collapsed building blocks in Figure~\ref{fig:rewrite}. 
This reveals a critical limitation in the generalization ability of existing models, which is closely tied to the ``input-dependency'' of editing instructions. For conventional explicit instructions (\eg, remove the object from the image), the correlation for instructions with the input image content is relatively low, like remove, which leads to consistent visual change logic across different objects and scenes (i.e., plausible removal of the target region and background completion). Thus, models can maintain good performance even on unseen scenes. In contrast, world knowledge-driven implicit instructions are highly input-contingent. As shown in Figure~\ref{intro}, applying the same action ``hit a cactus'' to balls of different materials results in vastly different visual outcomes (\eg, degree of deformation, reaction of the cactus) due to variations in physical properties (mass, elasticity, surface material, etc). This strong instruction–input–result coupling imposes far greater demands on model generalization than traditional tasks.

Thanks to the development of unified models~\citep{wang2024emu3,zhou2024transfusion,wang2025selftok,deng2025emerging}, which are based on large-scale pre-trained models capable of handling both comprehension and generation tasks, these models can not only rewrite editing instructions based on input images but, more importantly, they implicitly capture the causal logic and visual relationships underlying world editing instructions through exposure to large-scale pre-training data. They introduce a promising path toward world image editing: transferring the world knowledge embedded in unified models into the editing process. Thus, an image dataset specifically designed for world-driven editing to stimulate unified models to leverage their powerful reasoning capabilities for improving image editing is required. Although recent efforts such as AnyEdit~\citep{yu2025anyedit} have collected large-scale and diverse editing data, the vast majority still consist of traditional explicit instructions, with truly world knowledge-driven editing samples remaining scarce. KRIS-Bench~\citep{wu2025kris} and RISEBench~\citep{zhao2025envisioning} have proposed corresponding benchmark sets, but these cannot provide supervision for training and are limited in scale.

To address this gap, this paper proposes a comprehensive dataset \textbf{WorldEdit}. Using images from publicly available real-world segmentation datasets as original inputs, we employ instruction decomposition and multi-step editing via GPT-4o~\citep{hurst2024gpt} to generate edited images that conform to world knowledge logic. To ensure data quality, a two-stage filtering strategy—comprising instruction verification pre-filtering and image assessment post-filtering—is designed. The pre-filtering stage eliminates editing instructions with ambiguous causal logic through world knowledge consistency checks, while the post-filtering stage removes generated images that are visually unrealistic or violate world knowledge through visual plausibility assessment. Through this pipeline, the WorldEdit dataset ultimately contains 11k high-quality editing samples. 

Finally, to validate the effectiveness of WorldEdit, we conducted a two-stage training procedure based on Bagel~\citep{deng2025emerging}. In the first stage, we perform supervised fine-tuning with paraphrased instructions, enabling the model to better interpret implicit editing commands and align them with corresponding visual transformations. In the second stage, we further refine the model through reinforcement learning with a composite reward function that explicitly accounts for reasoning quality, visual fidelity, and causal consistency. This reward structure not only encourages the model to produce visually plausible and instruction-aligned images, but also grounds its generative process in interpretable reasoning traces and causal verification.
Our main contributions are summarized as follows:
\vspace{-0.5em}
\begin{itemize}[leftmargin=*, itemsep=0.3em, parsep=0.2em]
    \item We introduce \textbf{WorldEdit}, along with a challenging benchmark \textbf{WorldEdit-Test}, 
    specifically designed to capture \emph{cause-driven visual transformations} with world knowledge.

    \item With WorldEdit, we fine-tune Bagel in a two-stage process and introduce an 
    \emph{inversion-based causal verification reward} to better align generative behaviors with real-world causal logic.
    
    \item Our approach achieves \textbf{state-of-the-art} performance among open-source models on the WorldEdit-Test, demonstrating superior instruction generation and following capability.

\end{itemize}
\vspace{-0.5em}

\section{Related Work}

\subsection{Text-based Image Editing}

Image editing aims to modify image content according to given instructions while preserving both the consistency and naturalness of the edited output. Initially, diffusion models~\citep{rombach2022high,esser2024scaling}, due to their remarkable image generation capabilities, were adapted for image editing by altering diffusion trajectories~\citep{hertz2022prompt,tumanyan2023plug}. Subsequent research incorporating masks~\citep{avrahami2022blended,avrahami2023blended}, and multi-reference images~\citep{kumari2023multi,ruiz2023dreambooth} has significantly enhanced the controllability of image editing.
However, strong controllability does not necessarily imply intelligence. Since most diffusion-based methods rely on relatively small text encoders, they struggle to handle complex and fine-grained editing instructions that require reasoning.
Recently, researchers have begun to explore the use of unified models for image editing. These models extend the generative capabilities of large language models (LLMs) to the visual domain, enabling cross-modal generation and understanding. Recent approaches such as Chameleon~\citep{team2024chameleon}, Emu3~\citep{wang2024emu3}, and Selftok~\citep{wang2025selftok} adopt a unified next-token prediction paradigm by discretizing images, while Transfusion~\citep{zhou2024transfusion} and Show-o~\citep{xie2024show} integrate image diffusion with autoregressive text prediction within a single framework. Compared to diffusion-based approaches, these models demonstrate improved performance in instruction following and semantic understanding. Furthermore, several commercial systems, such as GPT-4o~\citep{hurst2024gpt} and Gemini-2.0-Flash~\citep{comanici2025gemini}, have exhibited impressive reasoning-based image editing capabilities, suggesting that unified LMMs provide a promising direction for image editing.
However, they still fall short when confronted with instructions that implicitly require world knowledge. This limitation arises because, although unified models possess strong comprehension and generation abilities, it remains an open question whether these abilities can be mutually reinforced, and how their pretrained knowledge and reasoning skills can be effectively leveraged to advance image editing.

\subsection{Benchmarks for Text-based Image Editing}
Collecting high-quality image editing data is inherently challenging, as it requires strong consistency between pre- and post-edit images. Recent datasets such as InstructPix2Pix~\citep{brooks2023instructpix2pix}, ImgEdit~\citep{ye2025imgedit}, AnyEdit~\citep{yu2025anyedit}, SEED-X~\citep{ge2024seed} and UniReal~\citep{chen2024unirealuniversalimagegeneration} employing strategies like synthetic generation and video sampling to curate large-scale, high-quality editing pairs. However, these efforts primarily emphasize task complexity, instruction diversity, or dataset scale, without explicitly modeling the reasoning processes or knowledge structures involved in instruction understanding.
More recent benchmark studies have begun to recognize this limitation by incorporating instruction understanding into their evaluations. For instance, SmartEdit~\citep{huang2024smartedit} investigates spatial and interactive reasoning in ambiguous editing scenarios. 
RISEBench~\citep{zhao2025envisioning} and KRIS-Bench~\citep{wu2025kris} support to evaluates knowledge reasoning in image editing, but they do not provide training data.
ReasonPix2Pix~\citep{jin2024reasonpix2pixinstructionreasoningdataset} and EditWorld~\citep{yang2024editworldsimulatingworlddynamics} only provided limited unverified knowledge editing data, and were constrained by early image editing techniques, resulting in poor image quality and insufficient knowledge.
Therefore, the lack of large-scale, high-quality training data continues to be a critical bottleneck.
In this paper, we introduce \textbf{WorldEdit}, a benchmark that addresses these gaps by providing both large-scale training and evaluation datasets. WorldEdit offers high-quality resources for supervised training of editing models, while enabling a more comprehensive assessment of their ability to integrate world knowledge, reasoning, and visual editing.

\section{WorldEdit}
\subsection{Construction of WorldEdit}
\textbf{Editing Type Definition.}
Most existing image editing benchmarks primarily focus on perceptual-level modifications or semantic operations, while often overlooking the rich spectrum of natural changes that occur in the real world. Previous efforts, such as RISE-Bench~\citep{zhao2025envisioning}, which organizes data according to types of cognitive reasoning, and KRIS-Bench~\citep{wu2025kris}, which categorizes tasks based on Bloom’s taxonomy of cognition, include a substantial portion of evaluation data centered on semantic manipulations (\eg, quantity, spatial location) and logical reasoning (\eg, mazes, Sudoku). As a result, data that simulate causal transformations in the physical world are systematically underrepresented in current benchmarks. To address this gap, we introduce \textbf{WorldEdit}, a benchmark comprising both training and test data. Our benchmark emphasizes how changes unfold under specific real-world constraints, with the goal of providing a more focused and comprehensive evaluation of models’ ability to reason about and simulate cause-driven visual transformations.

We categorize the transformations into two primary classes:
\textit{Environment-driven Transformations} result from changes in ambient conditions. Examples include Time (fruit ripening), Temperature (melting ice), Humidity (a flower wilting), Acidity (metal corrosion), and Light (fading fabric due to sunlight).
\textit{Mechanics-driven Transformations} result from the application of physical forces. Examples include Break (a plate shattering), Inflate (balloon expanding), Squeeze (paper crumpling), Twist (cloth wringing), and Stretch (rubber band elongating). A full definition of all transformation types is provided in the Appendix~\ref{Explanation}.

\begin{figure*}
  \centering
  \includegraphics[scale=0.58]{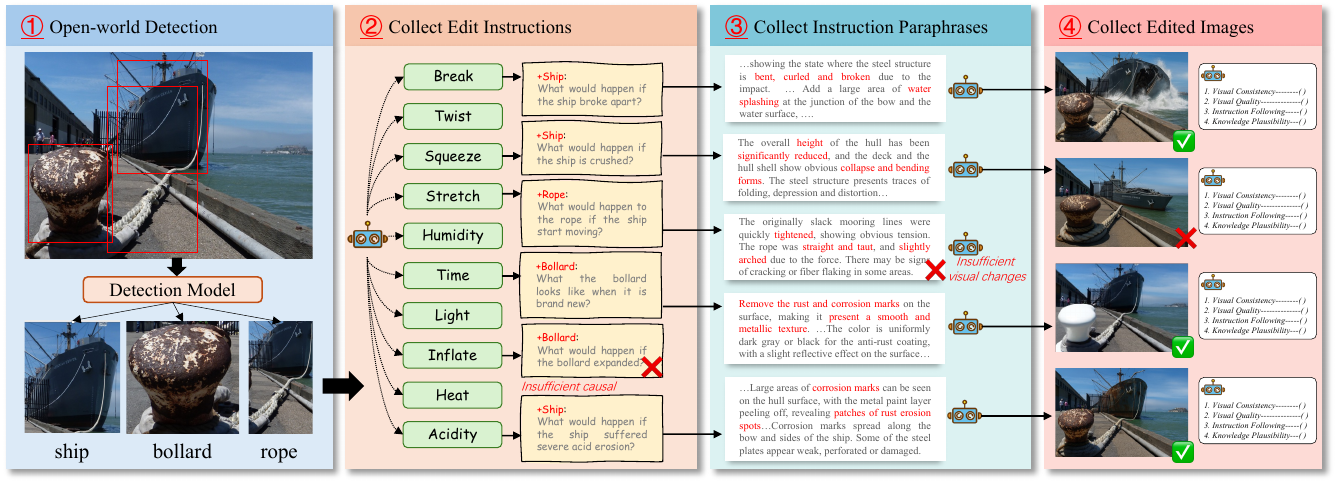}
\vspace{-1em}
\caption{The automated construction pipeline of the WorldEdit dataset. Open-world images are filtered and screened along three dimensions: (1) causal consistency of implicit instructions, (2) richness of the expected visual transformations, and (3) quality of the synthesized edited images.}
\label{pipeline}
\vspace{-1.5em}
\end{figure*}

\textbf{Editing Instructions, Explanations, and Images Collection.} 
To construct the WorldEdit, we design a pipeline, as shown in Figure~\ref{pipeline}, for collecting high-quality editing instructions, corresponding explanations, and edited images, ensuring coverage of the diverse real-world transformations.

First, we employ a detection model to extract object names from real-world images. This step aims to obtain the objects names and basic descriptions within the images, like \textit{a rusted bollard} and \textit{a white rope securing a ship to the bollard}. Accurate object detection is crucial as it forms the foundation for subsequent editing operations.

Next, we combine the detected objects with the predefined 10 editing types. For each object, we generate questions regarding its changes under different conditions. For example, for a ship, we might ask \textit{What would happen if the ship broke apart?}.
Notably, we perform filtering at this stage to remove unreasonable combinations or those with weak causal links. For example, the combination of \textit{Bollard} and \textit{Inflate} is filtered out because bollards, in reality, rarely undergo expansion, ensuring the collected data adheres to real-world plausibility.

We then utilize a pre-trained Large Language Model (LLM) to answer the valid questions obtained from the previous step. The LLM maps the causal changes arising from the questions to detailed visual change descriptions. 
Following this, we evaluate the generated visual change descriptions. We filter out responses with errors or those where the visual changes are not obvious. For instance, a description of \textit{a slightly taut rope} might be filtered as the resulting visual change is too subtle to be effectively edited and evaluated.

Finally, we leverage GPT-4o~\citep{hurst2024gpt} to generate edited images based on the paraphrased editing instructions. For more complex editing instructions (failed more than 3 times), we decompose them and performs multi-step editing to enhance the quality of the edited images. After image generation, GPT-4o is used again to evaluate the edited images. Images with poor visual consistency or incorrect edits are filtered out. For example, an image where a ship is edited to appear crushed but lacks convincing structural damage would be rejected. Through this comprehensive pipeline, we collect a large-scale, high-quality dataset that accurately reflects real-world causal visual transformations for the WorldEdit benchmark.

Although collecting data from the open-world images can cover most of the variations, considering the limitations of public data, some data, such as ``magnetic field lines'' and ``static effects'' are not common. Therefore, in order to increase the diversity of the data, we supplemented a portion of the synthetic data. For more details, please refer to the Appendix~\ref{Data_Collection}.

\begin{figure*}
  \centering
  \includegraphics[scale=0.4]{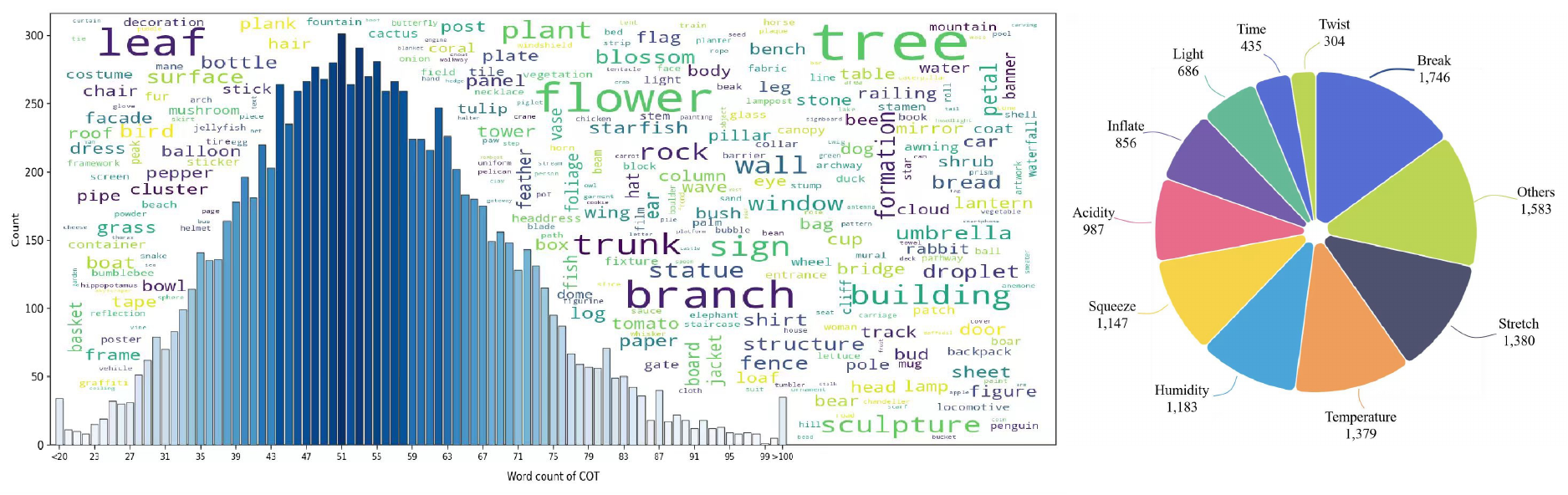}
\vspace{-1em}
\caption{Statistics of the WorldEdit dataset. (\textit{left}) Distribution of word counts in paraphrased instruction, along with a word cloud of frequently edited objects. (\textit{right}) Distribution of 10 editing instruction categories.}
\label{dataset}
\vspace{-2em}
\end{figure*}

\textbf{Dataset Characteristics and Statistics.}
Based on the data construction process, we have collected a total of 11k high-quality editing data. Among them, ``Break'' has the highest count (1,746), while ``Twist'' has the lowest count with only 304. We analyze it from multiple aspects, as in Figure~\ref{dataset}, objects like trees, flowers, and buildings are prominently featured, reflecting the dataset's wide coverage of both natural and man-made entities. The distribution of instruction explanation lengths is typically between 50 and 60 words, ensuring sufficient detail for understanding causal logic.

\subsection{Training with WorldEdit}

To assess the capability of WorldEdit in eliciting cross-modal reasoning and transferring knowledge from pre-trained multimodal models to visual generation tasks, we implemented a two-stage fine-tuning strategy using Bagel~\citep{deng2025emerging} as our baseline model. The initial stage involves supervised fine-tuning (SFT), where instructions are paraphrased into structured Chain-of-Thought (CoT) sequences. This phase aims to enhance the model’s capacity to interpret editing commands, leverage its inherent world knowledge, and establish a robust mapping from language instructions to visual modifications using a causal language modeling objective where loss is applied both to the text and image tokens.

Then, we employ reinforcement learning via the Flow-GRPO~\citep{liu2025flow} algorithm to further refine the model’s generative behavior. A composite reward function guides this optimization, integrating three complementary signals designed to ensure high-quality, reasoned, and causally consistent outputs.

\textbf{The CoT reasoning reward} is motivated by the need for transparent and logically sound internal reasoning. It evaluates the generated CoT text and outputs a scalar score, $R_{\text{reason}}$, based on the coherence, relevance, and correctness of the causal logic relative to the instruction. This reward encourages the model to produce rationales that faithfully reflect the transformation process, thereby improving interpretability and grounding the model’s decisions in its pre-existing knowledge.

\textbf{The image quality reward} ensures the visual output both aligns with the instruction and retains consistency with the input image. Taking as input the original image, the generated image, and the instruction, this reward uses a multimodal model to produce a score, $R_{\text{fidelity}}$, that captures instruction adherence and visual consistency. By directly optimizing the perceptual alignment and minimal invasiveness of the edit, this reward helps maintain high visual quality while executing precise changes.

\textbf{The causal verification reward} addresses the core objective of fostering genuine causal understanding. A multimodal model is required to infer the cause of transformation between the input and output images, and return a similarity score $R_{\text{causal}}$, between this inferred cause and the original instruction, as the reward. By incentivizing the model to produce edits that are not only visually correct but also causally explainable, this mechanism ensures that the model learns the underlying physical and environmental principles.

\begin{figure*}
  \centering
  \includegraphics[scale=0.42]{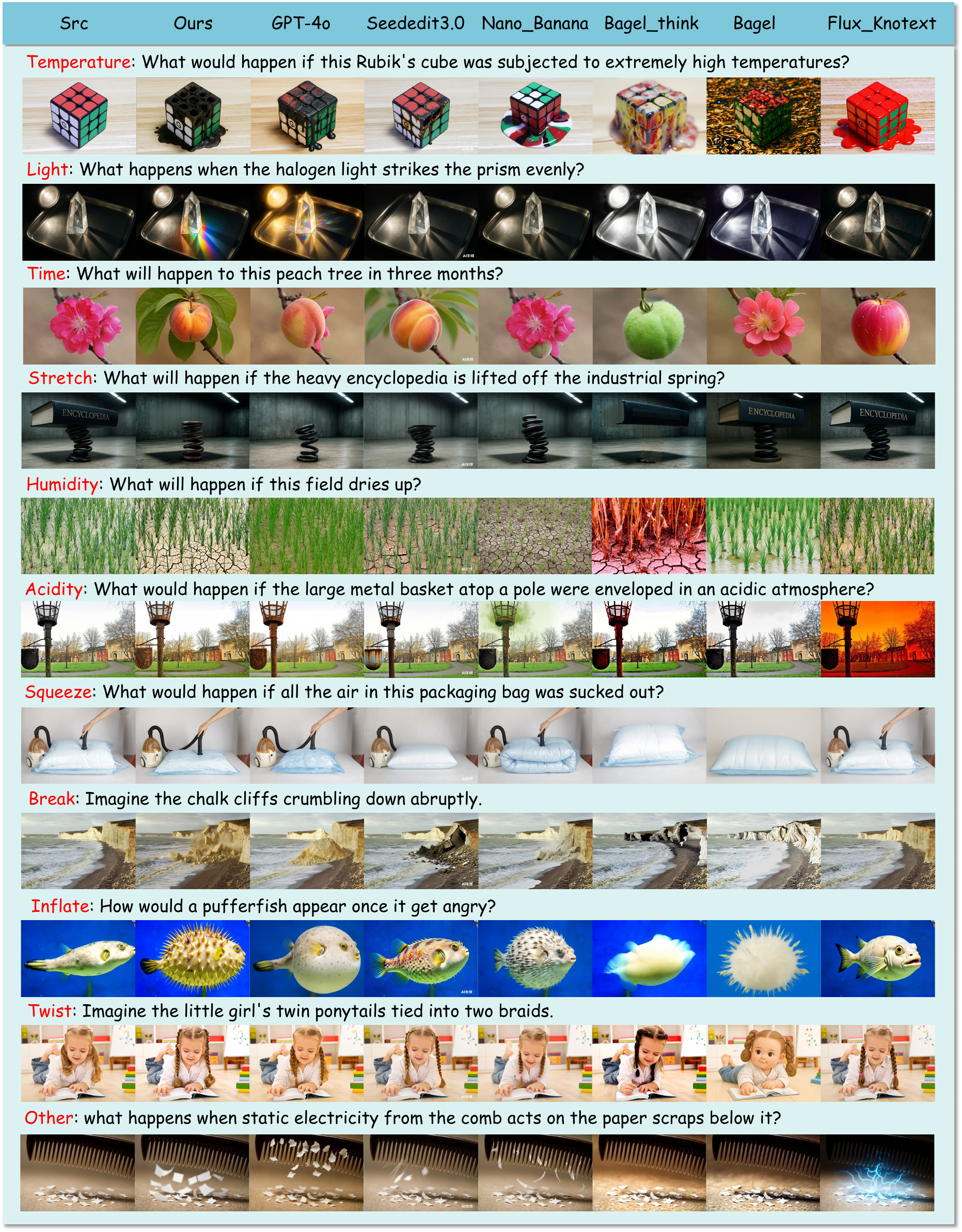}
\vspace{-0.5em}
\caption{Qualitative comparison across different causal categories. The figure shows representative examples from ten causal reasoning tasks.
Each row corresponds to a causal scenario, with the source image on the left followed by results from different models. 
Our method generates outputs that are both visually plausible and causally coherent, whereas baselines often produce irrelevant or stylistic edits, failing to reflect the causal logic of the instruction.
}
\label{fig:qualitative}
\vspace{-2.5em}
\end{figure*}

The overall reward is computed as $R = R_{\text{reason}} + R_{\text{fidelity}} + R_{\text{causal}}$. This multi-objective reward structure ensures the model is guided toward reasoning-aware, high-fidelity, and causally grounded image editing, effectively leveraging the rich cause-and-effect relationships embedded in the WorldEdit benchmark to support open-world visual transformation tasks.

\section{Experiments}

\begin{table}[]
\centering
\scalebox{0.5}{
\begin{tabular}{c|c|ccc|cccccccccc}
\toprule
\begin{tabular}[c]{@{}c@{}}Cause\\ Category\end{tabular} & Metric & GPT-4o  & Nano-Banana & SeedEdit-3.0& Ours & Flux-Kontext & Bagel-Think & Bagel  & Omnigen & Omnigen2 & Emu2 & Anyedit & Ip2p &Magicbrush \\
\midrule
\multirow{4}{*}{Time}                                         & VC                  &     4.02              &       4.22   & 3.81     &   4.18      &     4.6      &  2.38   &   2.34      &   3.02   &   2.22       &     1.48              &    1.63         &  3.04 &   1.40         \\
                                                              & VQ                  &     5.00 &5.00          &       4.96             &    4.55     &  4.88        & 4.52     &   4.02      &  4.20    &  4.35        &     4.56              &      3.25       &    4.51  &   4.28      \\
                                                              & IF                   &   3.86       &     3.90      & 3.64              &    3.71     &  1.46         &  2.46    &  1.66       &  1.38    &    1.33      &      1.54             &        1.79     &   1.51      & 1.19     \\
                                                              & KP                   & 4.20       &   4.14       &  3.79                & 3.94        &      1.42     &  2.86    &  1.80       &  1.48    &   1.25       &     1.75              &  1.92           &   1.65      &   1.34   \\
                                                              \cline{2-15}
                                                              & Avg                 &    \cellcolor{top2}{4.27}            &   \cellcolor{top1}{4.32}   &   4.05           &   \cellcolor{top3}{4.10}      &       3.09    &  3.06    &   2.46      & 2.52     &  2.29        &    2.33               &    2.15         & 2.68        & 2.05     \\
\midrule
\multirow{4}{*}{Temperature}                                  & VC                   & 4.02      &    4.22        &     3.98            &  3.52       &  3.86         &  2.08    &  1.94       &  2.54    & 2.26         &     1.38              &  1.45           &   2.62      &  1.81    \\
                                                              & VQ                   & 5.00      &   4.96        &  4.69                &   4.36      &   4.54        &   4.06   &  4.04       &  4.00    &  4.46        &    4.65               &   2.92          &   4.56      &  4.28    \\
                                                              & IF                   & 4.54     &   4.20        &    4.67               & 4.04        &     2.08      &  3.52    &  1.80       &  1.54    &     1.46     &     1.29              &          1.51   &   1.32     & 1.28      \\
                                                              & KP                   &    4.66 &       4.26         &    4.56           &  3.90       &       1.88    & 3.76     &   1.90      &  1.38    &   1.37       &   1.38                &  1.45           &  1.44       &1.30      \\
                                                              \cline{2-15}
                                                              & Avg                  & \cellcolor{top1}{4.56}      &   \cellcolor{top3}{4.41}         &     \cellcolor{top2}{4.47}            &  3.96       &       3.09    &  3.36    &   2.42      & 2.37     &  2.39        &   2.17                &         1.83    &  2.49       &  2.17    \\
\midrule
\multirow{4}{*}{Humidity}                                     & VC                   &  4.50    &    4.40        &  4.24                &   3.82      &   4.40        &  2.68    &   2.32      &  2.28    &  1.94        &      1.76             &    1.62         &  2.86     & 1.40       \\
                                                              & VQ                   &   4.94    &   4.90       &    4.76               &   4.58      &      4.94     & 4.42     & 4.32        &   4.14   &    4.50      &       4.56            &  3.52           & 4.62      &   4.28     \\
                                                              & IF                   &     3.92    &  3.52     &    4.22                &  3.28       &       1.52    &  3.36    & 2.22        &  1.34    &  1.54        &    1.88               &         1.90   &   1.72     &  1.36     \\
                                                              & KP                   &   4.26     &     3.76      &    4.38             &  3.44       &       1.60    &  3.36    &  2.16       & 1.44     &  1.64        &   2.20                &         2.06    & 1.90        & 1.40     \\
                                                              \cline{2-15}
                                                              & Avg                  &   \cellcolor{top1}{4.41}       &        \cellcolor{top3}{4.15}      &     \cellcolor{top2}{4.40}       &  3.78       &       3.12    &  3.46    &  2.76       &   2.30   &    2.41      &   2.60                &  2.28           & 2.78       &   2.11    \\
\midrule
\multirow{4}{*}{Acidity}                                      & VC                   &    4.46     &   4.36      &    4.14              &   3.94      &    4.74       &  2.74    &  2.60       &  3.14    &   1.88       &          1.34         &     1.38        & 2.74       &  1.60     \\
                                                              & VQ                   &    5.00  &    4.76         &      4.82           &   4.76      &   4.88        & 4.28     &   4.40      &   4.28   &  4.68        &         4.70          &          3.32   & 4.40    &  4.26       \\
                                                              & IF                  &  3.82    &    3.58        &  3.50                &     3.50    &     1.20      &   2.36   & 1.40        &  1.10    &      1.08    &    1.12               &  1.22           &  1.18    &  1.12       \\
                                                              & KP                   &      3.70 &    3.44        &  3.48               &  3.68       &     1.20      &  2.34    &   1.40      &  1.06    &   1.08       &      1.14             &  1.26           &  1.14       & 1.04     \\
                                                              \cline{2-15}
                                                              & Avg                  &    \cellcolor{top1}{4.25}  &   \cellcolor{top2}{4.04}          &     \cellcolor{top3}{3.99}            & 3.97        &       3.01    & 2.93     &   2.45      &  2.40    &   2.18       &     2.08              &        1.80     & 2.37      &   2.01     \\
\midrule
\multirow{4}{*}{Light}                                        & VC                   &   3.70     &    4.40       & 4.43                &  3.82       &   4.64        &   2.84   &  2.36       & 2.44     &   1.88       &      1.45             &    1.43         &   2.66      &  1.54    \\
                                                              & VQ                   &   4.78    &    4.92       &    4.86              & 4.68        &      4.84     &  4.46    &  4.26       &  4.26    &   4.48       &   4.71                &        2.98     &   4.50   &   4.20      \\
                                                              & IF                   &   4.40    &   2.88       &  3.74                 &  3.74       &    1.86       &  3.60    &  2.64       & 1.98     &    2.31      &  2.22                 &         2.11    &  1.62     &  1.41      \\
                                                              & KP                   &   4.64     &   3.50      &     4.06              &  4.12       &      2.22     & 3.84     &   3.00      & 2.20     &   2.67       &   2.67                &         2.28    &  2.06       &   1.83   \\
                                                              \cline{2-15}
                                                              & Avg                  &  \cellcolor{top1}{4.38} 
                                                              &  3.93          &      \cellcolor{top2}{4.27}              &     \cellcolor{top3}{4.09}    &        3.39   & 3.69     &  3.07       &  2.72    &     2.83     &     2.77              &         2.20    &  2.71      &  2.25     \\
\midrule
\multirow{4}{*}{Break}                                        & VC                   &  4.80     &    4.90        &     4.26            & 4.74        & 4.88          & 4.00     &  2.98       &2.98      & 1.80         &     1.36              &    1.46         &   2.36     &  1.40     \\
                                                              & VQ                   &   4.92     &  4.78       &  4.82                 &   4.48      &      4.72     & 4.12     &  4.06       & 4.54     &    4.46      &    4.64               &  3.28           &  4.50     & 4.14       \\
                                                              & IF                   & 4.06      &  4.34        & 3.98                  &   4.56      &     2.36      &  3.10    &  2.18       & 1.40     &    1.10      &     1.36              & 1.62            & 1.08     & 1.22        \\
                                                              & KP                   &    3.84   &   4.14         &     3.76            &  4.36       &        2.06   &  2.72    &   1.92      &  1.24    &     1.02     &    1.30               &       1.46      & 1.10        & 1.14     \\
                                                              \cline{2-15}
                                                              & Avg                  &   \cellcolor{top2}{4.41}      &     \cellcolor{top1}{4.54}      &    4.21            &  \cellcolor{top1}{4.54}       &       3.51    & 3.49     &  2.79       & 2.54     &    2.10      &   2.17                &        2.00     & 2.26       & 1.98      \\
\midrule
\multirow{4}{*}{Inflate}                                       & VC                   &   4.53      &  4.88       &   4.35               & 4.20        &   4.54        &  3.00    &  2.53       & 2.38     & 1.94         &      1.59             & 1.31            &  2.58      &  1.56     \\
                                                              & VQ                   &    4.94       &   4.98    &  4.86                &  4.46       &     4.92      & 4.46     &  4.12       & 3.98     &    4.47      & 4.57                  &        3.20     & 4.70       &  4.15     \\
                                                              & IF                   &    4.22    &   3.56      &    3.98               &  3.96       &      1.48     &  3.34    &    3.14     & 2.12     &    2.67      &   2.33                &  2.08           & 1.64     &   1.65      \\
                                                              & KP                   &   4.31    &   3.36       &  3.84                 & 3.70        &      1.46     & 3.06     &    2.94     &  1.78    &   2.37       &   2.37                &  1.74           &   1.5      &  1.60    \\
                                                              \cline{2-15}
                                                              & Avg                  &  \cellcolor{top1}{4.50}       &   \cellcolor{top3}{4.20}      &     \cellcolor{top2}{4.26}             &   4.08      &   3.10        & 3.47     &   3.18      & 2.57     &    2.86      &   2.71                &        2.08     &  2.61     &   2.24     \\
\midrule
\multirow{4}{*}{Squeeze}                                      & VC                  &  4.44     &     4.80        &     4.53           &   4.14      &  4.68         & 3.12     &  2.46       &   2.92   &   2.26       &       1.53            &   1.48          &   2.36     &  1.79     \\
                                                              & VQ                   &   4.94    &    4.94       & 4.66                 &  4.34       &    4.82       &  3.88    &   4.12      & 4.24     &   4.62       &     4.51              & 3.22            &  4.24      &  4.27     \\
                                                              & IF                   &   2.86             &  2.54    &     3.21         &  3.38       &       1.22    & 2.56     &   1.90      &  1.42    &   1.36       &     1.64              &         1.48    &  1.24     &   1.46     \\
                                                              & KP                   &    2.74 &   2.48         &     2.87              &  3.04       &        1.22   & 2.28     &  1.70       &   1.32   &     1.34     &      1.30             &  1.30           &  1.20       &   1.38   \\
                                                              \cline{2-15}
                                                              & Avg                  &   \cellcolor{top2}{3.75}       &   3.69    &    \cellcolor{top1}{3.82}               &   \cellcolor{top3}{3.73}      &       2.99    &  2.96    &  2.55       &  2.48    &    2.39      &   2.25                &  1.87           &   2.26    &   2.22     \\
\midrule
\multirow{4}{*}{Twist}                                        & VC                   &  4.52      &      4.52      &  4.45              & 4.3        &    4.52       &  2.56    &   2.22      &   2.62   &   2.02       &     1.63              &   1.43          &  1.84       &  1.66    \\
                                                              & VQ                   &   4.90   &    5.00        &  4.74                &  4.32       &     4.98      & 4.74     &  4.24       &  4.02    &   4.45       &    4.55               &        3.31     &  4.12     &  4.25      \\
                                                              & IF                   &    3.86     &   3.72     &     3.84              &  4.14       &        1.78   &  3.00    &   2.38      & 1.76     &   1.61       &   1.59                &         1.82    & 1.12    &   1.46       \\
                                                              & KP                   &    3.80    &   3.94      &     3.69              &   3.88      &        1.78   &   2.80   &  2.28       &  1.62    &   1.65       &   1.61                &   1.71          &  1.10       &   1.46   \\
                                                              \cline{2-15}
                                                              & Avg                 &  \cellcolor{top2}{4.27}  &  \cellcolor{top1}{4.30}            &    \cellcolor{top3}{ 4.18}             & 4.16        &       3.27    &  3.28    &  2.78       & 2.51     &     2.43     &    2.35               &  2.07           &   2.05     &  2.21     \\
\midrule
\multirow{4}{*}{Stretch}                                      & VC                 &   4.36     &    4.34        &  4.22              &  3.98       & 4.20          &  3.22    &   2.38      &  2.36    &  2.07        &        1.90           &    1.54         &   2.10      &  1.51   \\
                                                              & VQ                   &    5.00    &    4.98      & 4.78                 &   4.54      &       4.72    &4.52      &  3.98       &  4.14    &   4.54       &     4.59              &        3.10     &   4.10    & 3.79       \\
                                                              & IF                   &   4.44    &   3.84       &     3.88              &   3.96      &     2.00      & 3.44     &  2.60       & 2.18     &    2.33      &    2.55               &  2.06           &  1.08     &  1.04      \\
                                                              & KP                   &  4.52     &     3.76       & 4.04                &   4.10      &     2.12      &   3.48   &   2.50      &  2.10    &   2.48       &      2.71             & 2.04            &  1.08       &    1.21  \\
                                                              \cline{2-15}
                                                              & Avg                  &   \cellcolor{top1}{4.58}   &  \cellcolor{top2}{4.23}          &           \cellcolor{top2}{4.23}                &  4.15         &  3.26    &  3.67       & 2.87     &       2.70   &     2.85              &   2.94          &   2.19     & 2.09 &1.89     \\
\midrule
\multirow{4}{*}{Other}                                         & VC                   &  4.26       &     4.46       &4.52               &   4.22      &  3.96         & 3.20     & 2.68        & 2.20     &  3.54        &     2.10              &   2.08          &   2.50       &  1.86   \\
                                                              & VQ                   &    4.88    &     4.84      &  4.80               &     4.84    &    4.76       & 4.16     &  4.66       &  3.84    &   4.88       &      4.71             &         3.66    &   4.28     &  4.12     \\
                                                              & IF                   &    4.74 &   4.58         &     4.12              &  4.04       &       2.50    &  3.50    &  2.54       & 2.18     &    1.56      &  1.88                 &  2.24           & 1.74     &  1.80       \\
                                                              & KP                   &   4.68    &   4.56       &     4.10              &  3.96       &       2.70    &   3.38   &  2.52       &   2.10   &    1.56      &    1.90               &  2.20           &   1.72      &   1.96   \\
                                                              \cline{2-15}
                                                              & Avg                  &  \cellcolor{top1}{4.64}      &     \cellcolor{top2}{4.61}       &  \cellcolor{top3}{4.39}              &  4.27       &     3.48      &  3.56    & 3.10        & 2.58     &   2.89       &    2.65               &      2.55   & 2.56   &  2.44   \\
\midrule
Overall & Avg                  &    \cellcolor{top1}{4.36}             &  \cellcolor{top2}{4.22}    & \cellcolor{top3}{4.21}          & 4.07      &  3.21      &   3.35   &  2.76     & 2.52  &     2.51     &    2.46              & 2.09     &  2.44  &2.14  \\
\bottomrule
\end{tabular}
}
\vspace{-0.5em}
\caption{The performance of both commercial and open-source models on image editing tasks in WorldEdit-Test, evaluated across different causes and metrics. For each category, the \textcolor[rgb]{0.85,0.25,0.45}{Top-1}, \textcolor[rgb]{0.20,0.40,0.80}{Top-2}, and \textcolor[rgb]{0.85,0.65,0.10}{Top-3} scores are highlighted respectively.}
\label{tab:mian}
\vspace{-1.5em}
\end{table}

\subsection{Experimental Setup}
\textbf{Baselines.}  
We evaluate our method against a broad set of representative baselines, including traditional diffusion-based editing models and recent open-source unified multimodal models. In addition, we compare against commercial systems such as GPT-4o~\citep{hurst2024gpt}, SeedEdit-3.0~\citep{wang2025seededit}, and Nano-Banana~\citep{google2025gemini}, which have recently demonstrated impressive reasoning-aware editing ability.

\textbf{Metrics.}  
For evaluation, we follow the design of KRIS-Bench~\citep{wu2025kris} and adopt a multi-dimensional assessment protocol. Specifically, we employ Qwen-VL-Max~\citep{yang2025qwen3} as the evaluator to score editing results along four axes: visual consistency, which measures the structural alignment between the original and edited images; visual quality, which captures the perceptual realism of the generated images; instruction following, which quantifies how well the edits satisfy the given commands; and knowledge plausibility, which evaluates whether the edits are consistent with commonsense and world knowledge. More details are provided in the Appendix~\ref{Metrics}.  

\subsection{Main Results on WorldEdit-Test Benchmark}

\textbf{Compared with baselines.}
Table~\ref{tab:mian} reports the overall performance of different models across all causal categories and evaluation metrics. We observe that commercial models such as GPT-4o~\citep{hurst2024gpt} and SeedEdit-3.0~\citep{wang2025seededit} consistently excel in visual quality and instruction following, reflecting their advantages in large-scale pre-training and extensive instruction fine-tuning. However, despite their strong performance, they still face challenges in categories that require explicit causal reasoning, such as \emph{time} and \emph{acidity}.
Open-source editing models like Flux-Kontext~\citep{labs2025flux} are competitive in generating visually reasonable outputs but perform poorly in instruction following and knowledge consistency. This is particularly evident in categories such as "break", "squeeze", and "stretch", where capturing the causal dynamics of physical interactions is crucial. Similarly, open-source unified models outperform diffusion-based methods in instruction alignment but still have limitations in handling implicit instructions that require world knowledge.
Our method has made significant progress in knowledge consistency and instruction following, achieving state-of-the-art performance among all open-source models and competitive results compared to commercial systems.

\begin{figure*}
  \centering
  \includegraphics[scale=0.34]{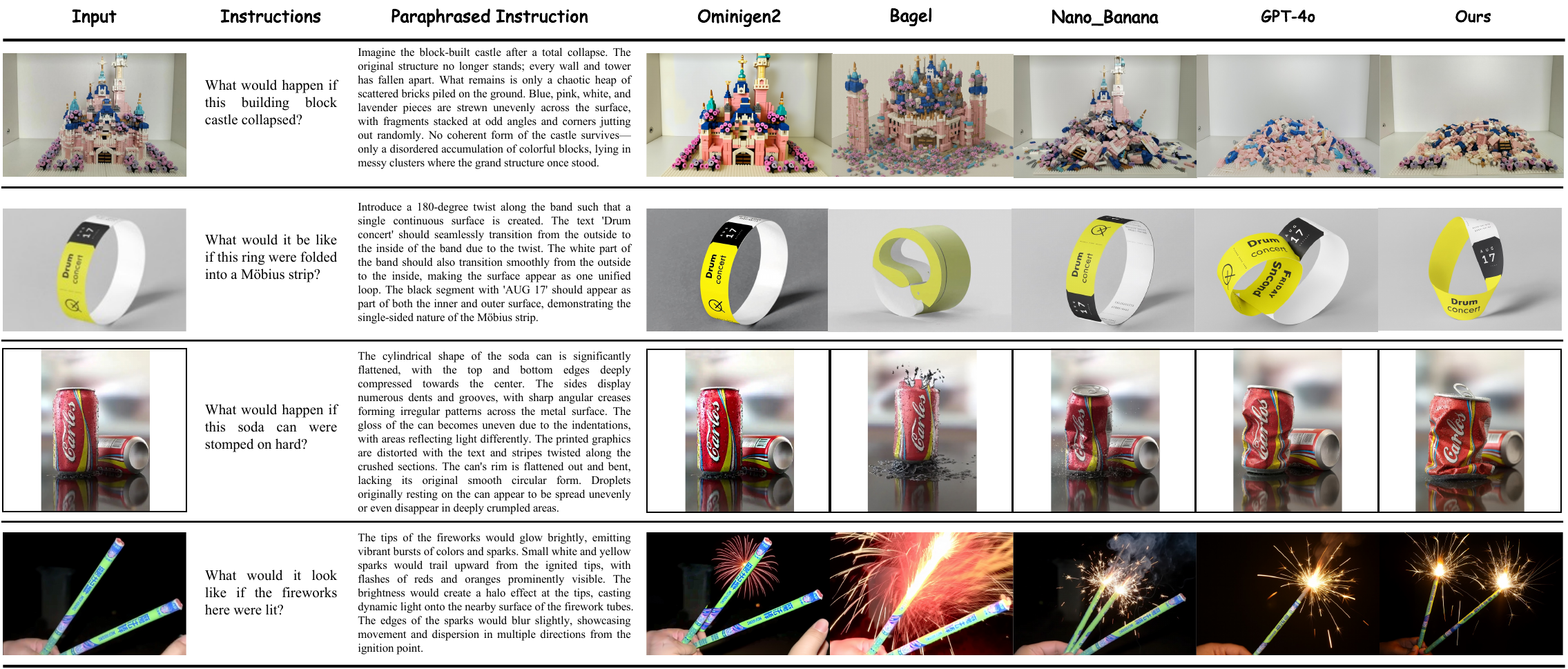}
\vspace{-1em}
\caption{Qualitative results on WorldEdit-Test with paraphrased instructions. Text alone often fails to capture fine-grained causal details (e.g., scattering pattern of collapsed building blocks), and models vary in their ability to interpret such prompts. Our model, fine-tuned with WorldEdit, generates the most faithful and visually coherent images, underscoring the importance of high-quality world knowledge-driven data.}
\label{fig:rewrite}
\vspace{-0.5em}
\end{figure*}

\begin{table}[]
\centering
\scalebox{0.68}{
\begin{tabular}{l|ccccccccccccccc}
\toprule
\begin{tabular}[c]{@{}c@{}}Methods\end{tabular} & Time   & Temperature & Humidity  & Acidity & Light & Break  & Inflate  & Squeeze  & Twist & Stretch & Other & Overall\\
\midrule
Nano-Banana$^{*}$                   & 4.35                    &  4.47             &   4.21                 & 4.10     & 3.93        & 4.55     & 4.21         &    3.70               &   4.37          & 4.26  &  4.63     & 4.25    \\
GPT-4o$^{*}$                      & 4.27             &  4.52     & 4.41              &   4.30                 &4.43      &   4.46      &  4.52    &  3.83        &     4.25              &   4.53          &    4.67   & 4.38       \\
\midrule
Omnigen2$^{*}$               & 2.44      & 2.40              & 2.70        &  2.30         & 2.92     &   2.36      &  2.84    &    2.19      &    2.42               &    2.87         &   3.33     & 2.62      \\
Bagel$^{*}$         & 3.39& 3.25&3.05 &2.53 &3.24 &2.75 &3.21 &2.46 & 2.37   &3.15 &3.71&3.01              \\

\midrule
Ours                    &   4.10           &3.96       &    3.78           & 3.97        &   4.09              &    4.54     & 4.08     &    3.73      &    4.16               & 4.15            &  4.27     & 4.07       \\
\bottomrule
\end{tabular}
}
\vspace{-0.5em}
\caption{Performance of different models on the WorldEdit-Test when using paraphrased instructions. Results are reported across ten causal categories and averaged overall.}
\label{tab:cot}
\vspace{-1.5em}
\end{table}

\textbf{Qualitative evaluation}
Figure~\ref{fig:qualitative} presents the visualization results under different causal categories. We can clearly observe the advantages of our method in terms of instruction following and knowledge plausibility. For example, in the ``temperature'' case (Row 1), our model successfully simulates the melting and deformation of the Rubik’s cube under high temperatures and reveals the internal bearing structure. In the ``light'' case, only our model and GPT-4o correctly decompose the light.
In categories involving strong physical interactions, such as ``stretch'', ``squeeze'', and ``break'' (Rows 4, 7, and 8), our method demonstrates outstanding capability in adhering to causal logic. Removing the heavy encyclopedia leads to a natural rebound of the spring, the vacuum-sealed bag collapses smoothly, and the cliffs break into visually coherent fragments—all consistent with the physical laws of the real world. In contrast, competing models often generate inconsistent or static results that fail to reflect the implied dynamic changes.

\textbf{Compared with paraphrased instruction.}
We further evaluated the model's editing results when directly given paraphrased instructions with clear visual changes. Our results reported in Table~\ref{tab:cot} indicate that although generating edited images based on paraphrased instructions results in improved accuracy across a majority of the tested scenarios, the performance gap remains significant when compared to the fine-tuning achieved with WorldEdit. For instance, while Bagel$^{*}$ shows a modest improvement of 0.25 over its original output, it still falls considerably short of the 4.07 achieved by our method. 
Moreover, as illustrated in Figure~\ref{fig:rewrite}, even detailed paraphrased instructions fail to enable the model to generate the lifelike results. In these complex cases, the paraphrased instructions alone are insufficient to convey the intricate details—such as the chaotic scattering of bricks in the castle or the abstract structure of the Möbius strip, which underscores the limitations of relying solely on textual adjustments. These finding highlights the critical role played by the specialized training data provided by WorldEdit in ensuring high-quality and accurate edits.

\subsection{Ablation Study}
To dissect the contributions of key components in our framework, we conducted ablation experiments on three reward functions in Table~\ref{tab:reward}.
Ablating the image quality reward, the overall score dropped by 0.22. Since image quality is a core requirement for visual editing, the absence of this reward to constrain the generation process led the model to produce less visually appealing outputs.
When the CoT reasoning reward and the causal verification reward are removed, the overall score decreases by 0.13 and 0.16 respectively. This result demonstrates that these two reward functions effectively constrain the model to maintain causal correctness during the editing process.

\begin{figure*}
  \centering
  \includegraphics[scale=0.36]{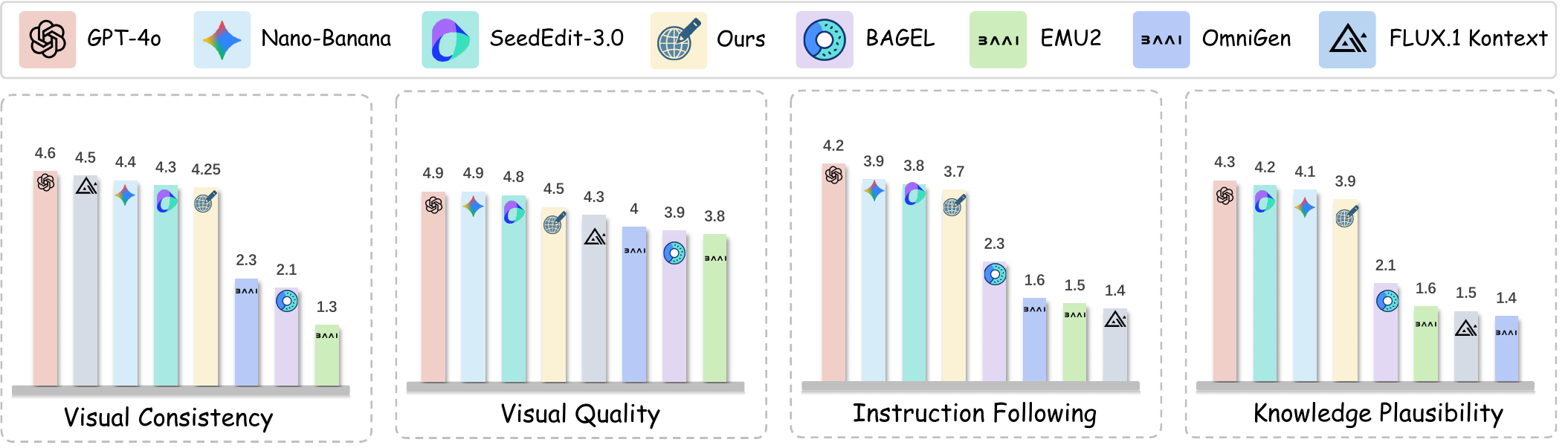}
\vspace{-1em}
\caption{Human evaluation results for various models on the WorldEdit-Test. GPT-4o consistently achieves the highest scores across all metrics. Ours shows solid performance, especially in instruction following and knowledge plausibility.}
\label{user}
\vspace{-1em}
\end{figure*}

\begin{table}[]
\centering
\scalebox{0.62}{
\begin{tabular}{l|ccccccccccccccc}
\toprule
\begin{tabular}[c]{@{}c@{}}Methods\end{tabular} & Time   & Temperature & Humidity  & Acidity & Light & Break  & Inflate  & Squeeze  & Twist & Stretch & Other & Overall\\
\midrule
w/o CoT reasoning reward  & 3.85  & 3.92        & 3.88     & 4.01    & 3.98   & 4.05   & 3.95    & 3.82    & 3.99    & 4.00    & 4.03    & 3.94    \\
w/o image quality reward  & 3.76  & 3.81        & 3.72     & 3.90    & 3.85   & 3.92   & 3.88    & 3.70    & 3.89    & 3.87    & 3.93    & 3.85    \\
w/o causal verification reward & 3.82  & 3.89        & 3.85     & 3.98    & 3.95   & 4.02   & 3.91    & 3.78    & 3.96    & 3.97    & 4.00    & 3.91    \\ 
\midrule
Ours &   4.10           &3.96       &    3.78           & 3.97        &   4.09              &    4.54     & 4.08     &    3.73      &    4.16               & 4.15            &  4.27     & 4.07       \\
\bottomrule
\end{tabular}
}
\vspace{-0.5em}
\caption{
Ablation study on the impact of different reward functions on the WorldEdit-Test.
}
\label{tab:reward}
\vspace{-1.5em}
\end{table}

\subsection{Human Evaluation}

From the human evaluation results in Figure~\ref{user}, several consistent conclusions emerge with Table~\ref{tab:mian}. Both evaluations highlight that GPT-4o and Nano-Banana lead in most categories, aligning with human judgment that these models exhibit superior capabilities in maintaining visual fidelity and adhering to instructions. Models like Flux-Kontext and Emu2 score significantly lower in knowledge plausibility, reflecting their struggles in executing real-world, physically consistent transformations.

\section{Conclusion}
In this paper, we presented WorldEdit, a novel large-scale dataset that emphasizes world-knowledge-driven image editing. By introducing a diverse set of causal transformations, we have enabled image editing models to go beyond simple attribute manipulation and engage in more complex, implicit reasoning tasks. WorldEdit not only provides a comprehensive collection of real-world scenarios for training but also introduces a benchmark, WorldEdit-Test, to assess the ability of models to generalize to causal scenarios.
Through the use of WorldEdit in two-stage fine-tuning and integration with causal verification rewards, we significantly improved Bagel's performance. The results align image edits with real-world causal logic, surpassing current state-of-the-art approaches.
Looking ahead, we aim to expand WorldEdit to include a wider range of causal scenarios, and we hope that WorldEdit can provide a crucial resource for advancing the field of knowledge-aware image editing, enabling the development of more sophisticated and autonomous systems in the future.

\bibliography{iclr2026_conference}

@inproceedings{linvinci,
  title={Vinci: Deep Thinking in Text-to-Image Generation using Unified Model with Reinforcement Learning},
  author={Lin, Wang and Hu, Wentao and Jia, Liyu and Pan, Kaihang and Majun, Zhang and Zhao, Zhou and Wu, Fei and Chen, Jingyuan and Zhang, Hanwang},
  booktitle={The Thirty-ninth Annual Conference on Neural Information Processing Systems},
  year={2025}
}

@inproceedings{rombach2022high,
  title={High-resolution image synthesis with latent diffusion models},
  author={Rombach, Robin and Blattmann, Andreas and Lorenz, Dominik and Esser, Patrick and Ommer, Bj{\"o}rn},
  booktitle={Proceedings of the IEEE/CVF conference on computer vision and pattern recognition},
  pages={10684--10695},
  year={2022}
}

@inproceedings{esser2024scaling,
  title={Scaling rectified flow transformers for high-resolution image synthesis},
  author={Esser, Patrick and Kulal, Sumith and Blattmann, Andreas and Entezari, Rahim and M{\"u}ller, Jonas and Saini, Harry and Levi, Yam and Lorenz, Dominik and Sauer, Axel and Boesel, Frederic and others},
  booktitle={Forty-first international conference on machine learning},
  year={2024}
}

@article{hertz2022prompt,
  title={Prompt-to-prompt image editing with cross attention control},
  author={Hertz, Amir and Mokady, Ron and Tenenbaum, Jay and Aberman, Kfir and Pritch, Yael and Cohen-Or, Daniel},
  journal={arXiv preprint arXiv:2208.01626},
  year={2022}
}

@inproceedings{tumanyan2023plug,
  title={Plug-and-play diffusion features for text-driven image-to-image translation},
  author={Tumanyan, Narek and Geyer, Michal and Bagon, Shai and Dekel, Tali},
  booktitle={Proceedings of the IEEE/CVF conference on computer vision and pattern recognition},
  pages={1921--1930},
  year={2023}
}

@inproceedings{avrahami2022blended,
  title={Blended diffusion for text-driven editing of natural images},
  author={Avrahami, Omri and Lischinski, Dani and Fried, Ohad},
  booktitle={Proceedings of the IEEE/CVF conference on computer vision and pattern recognition},
  pages={18208--18218},
  year={2022}
}

@article{avrahami2023blended,
  title={Blended latent diffusion},
  author={Avrahami, Omri and Fried, Ohad and Lischinski, Dani},
  journal={ACM transactions on graphics (TOG)},
  volume={42},
  number={4},
  pages={1--11},
  year={2023},
  publisher={ACM New York, NY, USA}
}

@article{team2024chameleon,
  title={Chameleon: Mixed-modal early-fusion foundation models},
  author={Team, Chameleon},
  journal={arXiv preprint arXiv:2405.09818},
  year={2024}
}

@article{wang2024emu3,
  title={Emu3: Next-token prediction is all you need},
  author={Wang, Xinlong and Zhang, Xiaosong and Luo, Zhengxiong and Sun, Quan and Cui, Yufeng and Wang, Jinsheng and Zhang, Fan and Wang, Yueze and Li, Zhen and Yu, Qiying and others},
  journal={arXiv preprint arXiv:2409.18869},
  year={2024}
}

@article{zhou2024transfusion,
  title={Transfusion: Predict the next token and diffuse images with one multi-modal model},
  author={Zhou, Chunting and Yu, Lili and Babu, Arun and Tirumala, Kushal and Yasunaga, Michihiro and Shamis, Leonid and Kahn, Jacob and Ma, Xuezhe and Zettlemoyer, Luke and Levy, Omer},
  journal={arXiv preprint arXiv:2408.11039},
  year={2024}
}

@article{xie2024show,
  title={Show-o: One single transformer to unify multimodal understanding and generation},
  author={Xie, Jinheng and Mao, Weijia and Bai, Zechen and Zhang, David Junhao and Wang, Weihao and Lin, Kevin Qinghong and Gu, Yuchao and Chen, Zhijie and Yang, Zhenheng and Shou, Mike Zheng},
  journal={arXiv preprint arXiv:2408.12528},
  year={2024}
}

@article{hurst2024gpt,
  title={Gpt-4o system card},
  author={Hurst, Aaron and Lerer, Adam and Goucher, Adam P and Perelman, Adam and Ramesh, Aditya and Clark, Aidan and Ostrow, AJ and Welihinda, Akila and Hayes, Alan and Radford, Alec and others},
  journal={arXiv preprint arXiv:2410.21276},
  year={2024}
}

@article{comanici2025gemini,
  title={Gemini 2.5: Pushing the frontier with advanced reasoning, multimodality, long context, and next generation agentic capabilities},
  author={Comanici, Gheorghe and Bieber, Eric and Schaekermann, Mike and Pasupat, Ice and Sachdeva, Noveen and Dhillon, Inderjit and Blistein, Marcel and Ram, Ori and Zhang, Dan and Rosen, Evan and others},
  journal={arXiv preprint arXiv:2507.06261},
  year={2025}
}

@inproceedings{brooks2023instructpix2pix,
  title={Instructpix2pix: Learning to follow image editing instructions},
  author={Brooks, Tim and Holynski, Aleksander and Efros, Alexei A},
  booktitle={Proceedings of the IEEE/CVF conference on computer vision and pattern recognition},
  pages={18392--18402},
  year={2023}
}

@article{ye2025imgedit,
  title={Imgedit: A unified image editing dataset and benchmark},
  author={Ye, Yang and He, Xianyi and Li, Zongjian and Lin, Bin and Yuan, Shenghai and Yan, Zhiyuan and Hou, Bohan and Yuan, Li},
  journal={arXiv preprint arXiv:2505.20275},
  year={2025}
}

@inproceedings{yu2025anyedit,
  title={Anyedit: Mastering unified high-quality image editing for any idea},
  author={Yu, Qifan and Chow, Wei and Yue, Zhongqi and Pan, Kaihang and Wu, Yang and Wan, Xiaoyang and Li, Juncheng and Tang, Siliang and Zhang, Hanwang and Zhuang, Yueting},
  booktitle={Proceedings of the Computer Vision and Pattern Recognition Conference},
  pages={26125--26135},
  year={2025}
}

@article{ge2024seed,
  title={Seed-x: Multimodal models with unified multi-granularity comprehension and generation},
  author={Ge, Yuying and Zhao, Sijie and Zhu, Jinguo and Ge, Yixiao and Yi, Kun and Song, Lin and Li, Chen and Ding, Xiaohan and Shan, Ying},
  journal={arXiv preprint arXiv:2404.14396},
  year={2024}
}

@inproceedings{huang2024smartedit,
  title={Smartedit: Exploring complex instruction-based image editing with multimodal large language models},
  author={Huang, Yuzhou and Xie, Liangbin and Wang, Xintao and Yuan, Ziyang and Cun, Xiaodong and Ge, Yixiao and Zhou, Jiantao and Dong, Chao and Huang, Rui and Zhang, Ruimao and others},
  booktitle={Proceedings of the IEEE/CVF Conference on Computer Vision and Pattern Recognition},
  pages={8362--8371},
  year={2024}
}

@article{zhao2025envisioning,
  title={Envisioning beyond the pixels: Benchmarking reasoning-informed visual editing},
  author={Zhao, Xiangyu and Zhang, Peiyuan and Tang, Kexian and Zhu, Xiaorong and Li, Hao and Chai, Wenhao and Zhang, Zicheng and Xia, Renqiu and Zhai, Guangtao and Yan, Junchi and others},
  journal={arXiv preprint arXiv:2504.02826},
  year={2025}
}

@article{wu2025kris,
  title={KRIS-Bench: Benchmarking Next-Level Intelligent Image Editing Models},
  author={Wu, Yongliang and Li, Zonghui and Hu, Xinting and Ye, Xinyu and Zeng, Xianfang and Yu, Gang and Zhu, Wenbo and Schiele, Bernt and Yang, Ming-Hsuan and Yang, Xu},
  journal={arXiv preprint arXiv:2505.16707},
  year={2025}
}

@inproceedings{kumari2023multi,
  title={Multi-concept customization of text-to-image diffusion},
  author={Kumari, Nupur and Zhang, Bingliang and Zhang, Richard and Shechtman, Eli and Zhu, Jun-Yan},
  booktitle={Proceedings of the IEEE/CVF conference on computer vision and pattern recognition},
  pages={1931--1941},
  year={2023}
}

@inproceedings{ruiz2023dreambooth,
  title={Dreambooth: Fine tuning text-to-image diffusion models for subject-driven generation},
  author={Ruiz, Nataniel and Li, Yuanzhen and Jampani, Varun and Pritch, Yael and Rubinstein, Michael and Aberman, Kfir},
  booktitle={Proceedings of the IEEE/CVF conference on computer vision and pattern recognition},
  pages={22500--22510},
  year={2023}
}

@InProceedings{Lim_2017_CVPR_Workshops,
  author = {Lim, Bee and Son, Sanghyun and Kim, Heewon and Nah, Seungjun and Lee, Kyoung Mu},
  title = {Enhanced Deep Residual Networks for Single Image Super-Resolution},
  booktitle = {The IEEE Conference on Computer Vision and Pattern Recognition (CVPR) Workshops},
  month = {July},
  year = {2017}
}

@article{deng2025emerging,
  title={Emerging properties in unified multimodal pretraining},
  author={Deng, Chaorui and Zhu, Deyao and Li, Kunchang and Gou, Chenhui and Li, Feng and Wang, Zeyu and Zhong, Shu and Yu, Weihao and Nie, Xiaonan and Song, Ziang and others},
  journal={arXiv preprint arXiv:2505.14683},
  year={2025}
}

@inproceedings{cao2023masactrl,
  title={Masactrl: Tuning-free mutual self-attention control for consistent image synthesis and editing},
  author={Cao, Mingdeng and Wang, Xintao and Qi, Zhongang and Shan, Ying and Qie, Xiaohu and Zheng, Yinqiang},
  booktitle={Proceedings of the IEEE/CVF international conference on computer vision},
  pages={22560--22570},
  year={2023}
}

@inproceedings{chung2024style,
  title={Style injection in diffusion: A training-free approach for adapting large-scale diffusion models for style transfer},
  author={Chung, Jiwoo and Hyun, Sangeek and Heo, Jae-Pil},
  booktitle={Proceedings of the IEEE/CVF conference on computer vision and pattern recognition},
  pages={8795--8805},
  year={2024}
}

@article{simsar2024uip2p,
  title={Uip2p: Unsupervised instruction-based image editing via cycle edit consistency},
  author={Simsar, Enis and Tonioni, Alessio and Xian, Yongqin and Hofmann, Thomas and Tombari, Federico},
  journal={arXiv preprint arXiv:2412.15216},
  year={2024}
}

@article{zhu2025kv,
  title={Kv-edit: Training-free image editing for precise background preservation},
  author={Zhu, Tianrui and Zhang, Shiyi and Shao, Jiawei and Tang, Yansong},
  journal={arXiv preprint arXiv:2502.17363},
  year={2025}
}

@inproceedings{simsar2025lime,
  title={Lime: localized image editing via attention regularization in diffusion models},
  author={Simsar, Enis and Tonioni, Alessio and Xian, Yongqin and Hofmann, Thomas and Tombari, Federico},
  booktitle={2025 IEEE/CVF Winter Conference on Applications of Computer Vision (WACV)},
  pages={222--231},
  year={2025},
  organization={IEEE}
}

@article{xu2023inversion,
  title={Inversion-free image editing with natural language},
  author={Xu, Sihan and Huang, Yidong and Pan, Jiayi and Ma, Ziqiao and Chai, Joyce},
  journal={arXiv preprint arXiv:2312.04965},
  year={2023}
}

@inproceedings{wang2023stylediffusion,
  title={Stylediffusion: Controllable disentangled style transfer via diffusion models},
  author={Wang, Zhizhong and Zhao, Lei and Xing, Wei},
  booktitle={Proceedings of the IEEE/CVF international conference on computer vision},
  pages={7677--7689},
  year={2023}
}

@inproceedings{yin2025grpose,
  title={Grpose: Learning graph relations for human image generation with pose priors},
  author={Yin, Xiangchen and Di, Donglin and Fan, Lei and Li, Hao and Chen, Wei and Song, Yang and Sun, Xiao and Yang, Xun and others},
  booktitle={Proceedings of the AAAI Conference on Artificial Intelligence},
  volume={39},
  number={9},
  pages={9526--9534},
  year={2025}
}

@article{shen2023advancing,
  title={Advancing pose-guided image synthesis with progressive conditional diffusion models},
  author={Shen, Fei and Ye, Hu and Zhang, Jun and Wang, Cong and Han, Xiao and Yang, Wei},
  journal={arXiv preprint arXiv:2310.06313},
  year={2023}
}

@article{wang2025seededit,
  title={SeedEdit 3.0: Fast and High-Quality Generative Image Editing},
  author={Wang, Peng and Shi, Yichun and Lian, Xiaochen and Zhai, Zhonghua and Xia, Xin and Xiao, Xuefeng and Huang, Weilin and Yang, Jianchao},
  journal={arXiv preprint arXiv:2506.05083},
  year={2025}
}

@article{yang2025qwen3,
  title={Qwen3 technical report},
  author={Yang, An and Li, Anfeng and Yang, Baosong and Zhang, Beichen and Hui, Binyuan and Zheng, Bo and Yu, Bowen and Gao, Chang and Huang, Chengen and Lv, Chenxu and others},
  journal={arXiv preprint arXiv:2505.09388},
  year={2025}
}

@misc{google2025gemini,
  title        = {Gemini Image Generation Overview},
  author       = {{DeepMind}},
  year         = {2025},
  howpublished = {\url{https://gemini.google/overview/image-generation/}},
  note         = {Accessed: 2025-09-25}
}

@article{liu2025flow,
  title={Flow-grpo: Training flow matching models via online rl},
  author={Liu, Jie and Liu, Gongye and Liang, Jiajun and Li, Yangguang and Liu, Jiaheng and Wang, Xintao and Wan, Pengfei and Zhang, Di and Ouyang, Wanli},
  journal={arXiv preprint arXiv:2505.05470},
  year={2025}
}

@article{labs2025flux,
  title={FLUX. 1 Kontext: Flow Matching for In-Context Image Generation and Editing in Latent Space},
  author={Labs, Black Forest and Batifol, Stephen and Blattmann, Andreas and Boesel, Frederic and Consul, Saksham and Diagne, Cyril and Dockhorn, Tim and English, Jack and English, Zion and Esser, Patrick and others},
  journal={arXiv preprint arXiv:2506.15742},
  year={2025}
}

@inproceedings{sun2024generative,
  title={Generative multimodal models are in-context learners},
  author={Sun, Quan and Cui, Yufeng and Zhang, Xiaosong and Zhang, Fan and Yu, Qiying and Wang, Yueze and Rao, Yongming and Liu, Jingjing and Huang, Tiejun and Wang, Xinlong},
  booktitle={Proceedings of the IEEE/CVF Conference on Computer Vision and Pattern Recognition},
  pages={14398--14409},
  year={2024}
}

@article{team2023gemini,
  title={Gemini: a family of highly capable multimodal models},
  author={Team, Gemini and Anil, Rohan and Borgeaud, Sebastian and Wu, Yonghui and Alayrac, Jean-Baptiste and Yu, Jiahui and Soricut, Radu and Schalkwyk, Johan and Dai, Andrew M and Hauth, Anja and others},
  journal={arXiv preprint arXiv:2312.11805},
  year={2023}
}

@article{liu2025step1x-edit,
      title={Step1X-Edit: A Practical Framework for General Image Editing},
      author={Shiyu Liu and Yucheng Han and Peng Xing and Fukun Yin and Rui Wang and Wei Cheng and Jiaqi Liao and Yingming Wang and Honghao Fu and Chunrui Han and Guopeng Li and Yuang Peng and Quan Sun and Jingwei Wu and Yan Cai and Zheng Ge and Ranchen Ming and Lei Xia and Xianfang Zeng and Yibo Zhu and Binxing Jiao and Xiangyu Zhang and Gang Yu and Daxin Jiang},
      journal={arXiv preprint arXiv:2504.17761},
      year={2025}
}

@article{xiao2024omnigen,
  title={Omnigen: Unified image generation},
  author={Xiao, Shitao and Wang, Yueze and Zhou, Junjie and Yuan, Huaying and Xing, Xingrun and Yan, Ruiran and Li, Chaofan and Wang, Shuting and Huang, Tiejun and Liu, Zheng},
  journal={arXiv preprint arXiv:2409.11340},
  year={2024}
}

@misc{2025hidream,
    title={HiDream-I1},
    author={HiDream.ai},
    howpublished = {\url{https://github.com/HiDream-ai/HiDream-I1}},
    year={2025}
}

@misc{flux2024,
    author={Black Forest Labs},
    title={FLUX},
    year={2024},
    howpublished={\url{https://github.com/black-forest-labs/flux}},
}

@misc{chen2024unirealuniversalimagegeneration,
      title={UniReal: Universal Image Generation and Editing via Learning Real-world Dynamics}, 
      author={Xi Chen and Zhifei Zhang and He Zhang and Yuqian Zhou and Soo Ye Kim and Qing Liu and Yijun Li and Jianming Zhang and Nanxuan Zhao and Yilin Wang and Hui Ding and Zhe Lin and Hengshuang Zhao},
      year={2024},
      eprint={2412.07774},
      archivePrefix={arXiv},
      primaryClass={cs.CV},
      url={https://arxiv.org/abs/2412.07774}, 
}

@misc{yang2024editworldsimulatingworlddynamics,
      title={EditWorld: Simulating World Dynamics for Instruction-Following Image Editing}, 
      author={Ling Yang and Bohan Zeng and Jiaming Liu and Hong Li and Minghao Xu and Wentao Zhang and Shuicheng Yan},
      year={2024},
      eprint={2405.14785},
      archivePrefix={arXiv},
      primaryClass={cs.CV},
      url={https://arxiv.org/abs/2405.14785}, 
}

@misc{jin2024reasonpix2pixinstructionreasoningdataset,
      title={ReasonPix2Pix: Instruction Reasoning Dataset for Advanced Image Editing}, 
      author={Ying Jin and Pengyang Ling and Xiaoyi Dong and Pan Zhang and Jiaqi Wang and Dahua Lin},
      year={2024},
      eprint={2405.11190},
      archivePrefix={arXiv},
      primaryClass={cs.CV},
      url={https://arxiv.org/abs/2405.11190}, 
}

@inproceedings{emotion,
  title={Exploring Embodied Emotion Through A Large-Scale Egocentric Video Dataset},
  author={Lin, Wang and Feng, Yueying and Han, WenKang and Jin, Tao and Zhao, Zhou and Wu, Fei and Yao, Chang and Chen, Jingyuan},
  booktitle={The Thirty-eight Conference on Neural Information Processing Systems Datasets and Benchmarks Track}
}

@article{lin2024non,
  title={Non-confusing Generation of Customized Concepts in Diffusion Models},
  author={Lin, Wang and Chen, Jingyuan and Shi, Jiaxin and Zhu, Yichen and Liang, Chen and Miao, Junzhong and Jin, Tao and Zhao, Zhou and Wu, Fei and Yan, Shuicheng and others},
  journal={arXiv preprint arXiv:2405.06914},
  year={2024}
}

@inproceedings{linaction,
  title={Action Imitation in Common Action Space for Customized Action Image Synthesis},
  author={Lin, Wang and Chen, Jingyuan and Shi, Jiaxin and Guo, Zirun and Zhu, Yichen and Wang, Zehan and Jin, Tao and Zhao, Zhou and Wu, Fei and Shuicheng, YAN and others},
  booktitle={The Thirty-eighth Annual Conference on Neural Information Processing Systems}
}

@article{wang2025towards,
  title={Towards transformer-based aligned generation with self-coherence guidance},
  author={Wang, Shulei and Lin, Wang and Huang, Hai and Wang, Hanting and Cai, Sihang and Han, WenKang and Jin, Tao and Chen, Jingyuan and Sun, Jiacheng and Zhu, Jieming and others},
  journal={arXiv preprint arXiv:2503.17675},
  year={2025}
}

@inproceedings{lin2023tavt,
  title={TAVT: Towards Transferable Audio-Visual Text Generation},
  author={Lin, Wang and Jin, Tao and Pan, Wenwen and Li, Linjun and Cheng, Xize and Wang, Ye and Zhao, Zhou},
  booktitle={Proceedings of the 61st Annual Meeting of the Association for Computational Linguistics (Volume 1: Long Papers)},
  pages={14983--14999},
  year={2023}
}

@inproceedings{lin2023exploring,
  title={Exploring group video captioning with efficient relational approximation},
  author={Lin, Wang and Jin, Tao and Wang, Ye and Pan, Wenwen and Li, Linjun and Cheng, Xize and Zhao, Zhou},
  booktitle={Proceedings of the IEEE/CVF International Conference on Computer Vision},
  pages={15281--15290},
  year={2023}
}

@inproceedings{pan2025generative,
  title={Generative Multimodal Pretraining with Discrete Diffusion Timestep Tokens},
  author={Pan, Kaihang and Lin, Wang and Yue, Zhongqi and Ao, Tenglong and Jia, Liyu and Zhao, Wei and Li, Juncheng and Tang, Siliang and Zhang, Hanwang},
  booktitle={Proceedings of the Computer Vision and Pattern Recognition Conference},
  pages={26136--26146},
  year={2025}
}

@article{wang2025selftok,
  title={Selftok: Discrete Visual Tokens of Autoregression, by Diffusion, and for Reasoning},
  author={Wang, Bohan and Yue, Zhongqi and Zhang, Fengda and Chen, Shuo and Bi, Li'an and Zhang, Junzhe and Song, Xue and Chan, Kennard Yanting and Pan, Jiachun and Wu, Weijia and others},
  journal={arXiv preprint arXiv:2505.07538},
  year={2025}
}

@article{li2023multi,
  title={Multi-granularity relational attention network for audio-visual question answering},
  author={Li, Linjun and Jin, Tao and Lin, Wang and Jiang, Hao and Pan, Wenwen and Wang, Jian and Xiao, Shuwen and Xia, Yan and Jiang, Weihao and Zhao, Zhou},
  journal={IEEE Transactions on Circuits and Systems for Video Technology},
  year={2023},
  publisher={IEEE}
}

@inproceedings{wang2023weakly,
  title={Weakly-supervised spoken video grounding via semantic interaction learning},
  author={Wang, Ye and Lin, Wang and Zhang, Shengyu and Jin, Tao and Li, Linjun and Cheng, Xize and Zhao, Zhou},
  booktitle={Proceedings of the 61st Annual Meeting of the Association for Computational Linguistics (Volume 1: Long Papers)},
  pages={10914--10932},
  year={2023}
}

@inproceedings{wang2023semantic,
  title={Semantic-conditioned dual adaptation for cross-domain query-based visual segmentation},
  author={Wang, Ye and Jin, Tao and Lin, Wang and Cheng, Xize and Li, Linjun and Zhao, Zhou},
  booktitle={Findings of the Association for Computational Linguistics: ACL 2023},
  pages={9797--9815},
  year={2023}
}

@inproceedings{guo2025bridging,
  title={Bridging the gap for test-time multimodal sentiment analysis},
  author={Guo, Zirun and Jin, Tao and Xu, Wenlong and Lin, Wang and Wu, Yangyang},
  booktitle={Proceedings of the AAAI Conference on Artificial Intelligence},
  volume={39},
  number={16},
  pages={16987--16995},
  year={2025}
}

@inproceedings{yan2024low,
  title={Low-rank Prompt Interaction for Continual Vision-Language Retrieval},
  author={Yan, Weicai and Wang, Ye and Lin, Wang and Guo, Zirun and Zhao, Zhou and Jin, Tao},
  booktitle={Proceedings of the 32nd ACM International Conference on Multimedia},
  pages={8257--8266},
  year={2024}
}

@inproceedings{yan2025diff,
  title={Diff-prompt: Diffusion-driven prompt generator with mask supervision},
  author={Yan, Weicai and Lin, Wang and Guo, Zirun and Wang, Ye and Feng, Fangming and Yang, Xiaoda and Wang, Zehan and Jin, Tao},
  booktitle={The Thirteenth International Conference on Learning Representations},
  year={2025}
}

@article{guo2025efficient,
  title={Efficient prompting for continual adaptation to missing modalities},
  author={Guo, Zirun and Wang, Shulei and Lin, Wang and Yan, Weicai and Wu, Yangyang and Jin, Tao},
  journal={arXiv preprint arXiv:2503.00528},
  year={2025}
}

@inproceedings{jin2024rethinking,
  title={Rethinking the multimodal correlation of multimodal sequential learning via generalizable attentional results alignment},
  author={Jin, Tao and Lin, Wang and Wang, Ye and Li, Linjun and Cheng, Xize and Zhao, Zhou},
  booktitle={Proceedings of the 62nd Annual Meeting of the Association for Computational Linguistics (Volume 1: Long Papers)},
  pages={5247--5265},
  year={2024}
}

@article{jin2025recognize,
  title={Recognize-and-tell: Generating video captions with textual cue in scene},
  author={Jin, Tao and Lin, Wang and Jiang, Hao and Wang, Jian and Jin, Xiao and Zhang, Zhimeng and Chen, Jingyuan and Zhao, Zhou and Zhang, Zhongfei},
  journal={Expert Systems with Applications},
  pages={127831},
  year={2025},
  publisher={Elsevier}
}

@article{lin2025reasoning,
  title={Reasoning physical video generation with diffusion timestep tokens via reinforcement learning},
  author={Lin, Wang and Jia, Liyu and Hu, Wentao and Pan, Kaihang and Yue, Zhongqi and Zhao, Wei and Chen, Jingyuan and Wu, Fei and Zhang, Hanwang},
  journal={arXiv preprint arXiv:2504.15932},
  year={2025}
}

@article{wang2025omni,
  title={Omni-Chart-600K: A Comprehensive Dataset of Chart Types for Chart Understanding},
  author={Wang, Shulei and Yang, Shuai and Lin, Wang and Guo, Zirun and Cai, Sihang and Huang, Hai and Wang, Ye and Chen, Jingyuan and Jin, Tao},
  journal={Findings of the Association for Computational Linguistics: NAACL 2025},
  pages={4051--4069},
  year={2025}
}

@inproceedings{lin2025non,
  title={Non-Natural Image Understanding with Advancing Frequency-based Vision Encoders},
  author={Lin, Wang and Wang, QingSong and Feng, Yueying and Wang, Shulei and Jin, Tao and Zhao, Zhou and Wu, Fei and Yao, Chang and Chen, Jingyuan},
  booktitle={Proceedings of the Computer Vision and Pattern Recognition Conference},
  pages={29756--29766},
  year={2025}
}

@article{huang2024autogeo,
  title={Autogeo: Automating geometric image dataset creation for enhanced geometry understanding},
  author={Huang, Zihan and Wu, Tao and Lin, Wang and Zhang, Shengyu and Chen, Jingyuan and Wu, Fei},
  journal={arXiv preprint arXiv:2409.09039},
  year={2024}
}

@inproceedings{wu2024semantic,
  title={Semantic Alignment for Multimodal Large Language Models},
  author={Wu, Tao and Li, Mengze and Chen, Jingyuan and Ji, Wei and Lin, Wang and Gao, Jinyang and Kuang, Kun and Zhao, Zhou and Wu, Fei},
  booktitle={Proceedings of the 32nd ACM International Conference on Multimedia},
  pages={3489--3498},
  year={2024}
}

@inproceedings{wang2024instruction,
  title={Instruction Tuning-free Visual Token Complement for Multimodal LLMs},
  author={Wang, Dongsheng and Cui, Jiequan and Li, Miaoge and Lin, Wang and Chen, Bo and Zhang, Hanwang},
  booktitle={European Conference on Computer Vision},
  pages={446--462},
  year={2024},
  organization={Springer}
}
\bibliographystyle{iclr2026_conference}

\appendix

\section{Detailed Tasks Explanation}
\label{Explanation}
In this section, we provide detailed explanations of the editing tasks used in the WorldEdit benchmark. 
Each task type corresponds to a specific causal transformation mode, covering both environment-driven 
and mechanics-driven changes, as well as an ``Other'' category for phenomena not captured by the 
ten canonical modes. Together, these modes ensure comprehensive coverage of real-world 
knowledge-informed editing scenarios.

\textbf{Time.} 
This mode refers to visual transformations that unfold naturally as time progresses, driven by 
biological growth, chemical reactions, or material aging. Typical examples include fruit ripening, 
flowers wilting, food decaying, or seasonal changes in plants. Beyond these, the mode also covers 
oxidation effects such as a bitten apple surface turning brown, structural aging such as buildings 
weathering and deteriorating, and life cycle developments such as tadpoles growing into frogs or 
bamboo shoots maturing into bamboo stalks. Edits under this mode simulate how objects evolve 
as time passes, often involving color fading, structural growth, surface corrosion, or progressive 
degradation consistent with real-world temporal dynamics.

\textbf{Temperature.} 
This mode captures a wide range of transformations driven by thermal conditions, 
including heating, ignition, burning, melting, and freezing. Classic examples include ice melting 
under heat, chocolate softening and dripping, snowmen collapsing as they thaw, or plastic deforming 
and liquefying at high temperatures. Different materials display distinct visual and textural changes: 
iron glows red and emits light when exposed to several hundred degrees of heat, while fireworks 
burst into sparks and colorful trails upon ignition. Cooling or freezing leads to effects such as 
solidification, frost formation, or stiffness in meat and other organic matter. This mode emphasizes 
how varying temperature conditions trigger material-specific responses, resulting in visually diverse outcomes across different substances.

\textbf{Humidity.} 
This mode concerns transformations caused by the absorption or loss of moisture. 
Representative cases include dry noodles softening when soaked, food becoming moldy, or plants 
wilting when deprived of water. Beyond these, it also encompasses material- and environment-driven 
changes such as sponges swelling and softening after absorbing water, dried fungi like black fungus 
or seaweed shrinking and hardening after dehydration, floors appearing darker and glossier when wet, 
and moss growing on damp stones. Biological cases are also included, such as animal fur becoming 
flattened and clumped when soaked. Edits in this mode emphasize how the presence or absence of 
moisture alters texture, volume, and surface appearance, capturing both swelling and shrinking 
effects consistent with real-world humidity dynamics.

\textbf{Acidity.} 
This mode represents transformations driven by acidic environments, where chemical 
erosion gradually degrades materials or organisms. Classical cases include metal rusting, paint 
peeling, or surfaces developing corrosion stains. Beyond these, acidity can also affect organic 
objects such as plants exposed to acid rain, leading to leaf damage and decay, or fabrics and 
clothing that lose color, develop holes, or weaken under acidic exposure. Edits in this mode 
emphasize chemical wear-and-tear, manifested through discoloration, surface roughness, 
structural weakening, and progressive material breakdown consistent with corrosive processes.

\textbf{Light.} 
This mode includes transformations driven by illumination conditions, where changes in the 
direction, intensity, or nature of light lead to distinct visual outcomes. Common examples are the 
dispersion of light through prisms, the focusing and diverging of light beams, or shifted shadows cast 
by different light sources. The mode also covers scenarios such as changes in shadow shapes and lengths due to altered light positions, ultraviolet exposure affecting surfaces, transitions 
between near-beam and far-beam headlights, or adding a specific spotlight to highlight part of a scene. Edits in this mode emphasize optical phenomena such as refraction, concentration, and shadow dynamics, as well as the role of light strength in altering the visual atmosphere of the scene.

\textbf{Break.} 
This mode describes destructive transformations where an object undergoes structural 
failure once its internal strength is exceeded. Such changes may be driven by \textit{external forces}, 
such as impacts, collisions, or pulling and tearing, leading to effects like shattering glass, collapsing 
cliffs, or ripping fabric. Alternatively, they may arise from \textit{internal stresses} accumulated within 
the object, such as tree branches snapping under their own weight, ceramics cracking from uneven 
thermal expansion, or water freezing inside a container and causing rupture. Edits in this mode 
simulate cracks, fractures, splinters, or complete disintegration, reflecting real-world principles of 
mechanical instability under external pressure or internal stress.

\textbf{Inflate.} 
This mode characterizes transformations in which an object becomes larger, fuller, or more 
voluminous due to internal growth, external force, or energy accumulation. This includes classical 
cases such as a balloon being inflated, a pufferfish puffing up, or dough rising. Beyond these, the 
mode also encompasses natural phenomena where entities expand in scale or density, such as 
trees growing more luxuriant, water flows becoming greater and more turbulent, or clouds swelling 
and spreading across the sky. Edits in this mode emphasize visual cues of enlargement, increased 
density, and dynamic expansion, while maintaining coherence with the object's inherent structure 
and properties.

\textbf{Squeeze.} 
This mode represents deformation caused by external compression, collision, or the loss 
of structural support. Typical examples include crumpled paper, a soda can stomped flat, or air 
being sucked out of a plastic bag. Beyond these, it also covers larger-scale or organic cases such as 
a car body being deformed after a rear-end collision, or skin forming wrinkles under pressure 
. Edits in this mode highlight flattened or distorted shapes, visible creases, dents, and surface 
irregularities, reflecting how objects lose volume or structure when squeezed or compressed.

\textbf{Twist.} 
This mode denotes transformations where objects deviate from their original orderly or 
linear structure into bent, entangled, or irregular configurations. Unlike purely torsional forces that 
produce simple spiral shapes, this mode emphasizes more general distortions involving bending, 
wrapping, or chaotic growth. Representative examples include folding a paper strip into a Möbius 
band, tangled earphone wires, tree branches growing in irregular directions, or a straight iron rod 
being arbitrarily twisted multiple times. Edits in this mode highlight irregular curvatures, interwoven 
structures, and complex spatial rearrangements while preserving continuity with the original object.

\textbf{Stretch.} 
This mode corresponds to transformations where objects are elongated, straightened, or 
expanded under tensile force or external manipulation. Typical examples include a rubber band being 
pulled, dough stretched thin, or springs extended. Beyond these, the mode also covers cases where 
initially compact, wrinkled, or bent structures are unfolded or spread out. Representative examples 
include tangled earphone wires being straightened, plants growing upward and outward, a sweater 
being pulled outward with its knitted patterns extended accordingly, wrinkled clothes being smoothed 
flat, animals spreading their wings, or a fishing net originally clustered together being stretched open. 
Edits in this mode emphasize proportional lengthening, surface expansion, or the transition from 
compactness to openness, while preserving material continuity and structural coherence.

\textbf{Other.} 
This category encompasses causal transformations not fully captured by the ten predefined 
modes, while remaining consistent with real-world physical or environmental principles. 
Representative cases include electrostatic attraction of objects, magnetic field interactions. It also covers 
material and fluid phenomena like sedimentation in liquids, diffusion and dispersion processes, or sudden events like a balloon bursting. In addition, broader contextual or scene-level changes, such as festive transformations involving decorations, lighting, or fireworks, are also included. 
This category ensures flexibility for capturing uncommon but physically valid effects beyond the above modes.

\section{More Details about Data Collection}
\label{Data_Collection}

To construct the WorldEdit dataset, we developed a multi-stage pipeline designed to ensure both 
instructional diversity and visual plausibility of the edited results. Compared with existing 
benchmarks such as KRIS-Bench~\citep{wu2025kris} and RISE-Bench~\citep{zhao2025envisioning}, our construction process emphasizes causal 
transformations grounded in world knowledge and integrates both automated and human-in-the-loop 
filtering mechanisms. Below, we provide further details of each stage.

\textbf{Data Sources.} 
The foundation of our dataset is the DF2K-OST~\citep{Lim_2017_CVPR_Workshops} collection, which provides high-resolution natural 
images across diverse domains. To emphasize realistic and visually rich scenarios, we filtered out 
all low-resolution samples and retained only high-quality images. Each image was then center-cropped 
into $2{:}3$ and $3{:}2$ aspect ratios to standardize the format. Since cropping may occasionally 
truncate essential semantic content, we employed GPT-4o~\citep{hurst2024gpt} to automatically filter out images that 
lost contextual coherence after cropping. The remaining subset was taken as the primary pool of 
original images. To further enrich causal diversity, we supplemented the dataset with additional 
images curated from the internet that emphasize strong logical and causal relationships. Moreover, 
synthetic samples generated with GPT-4o were incorporated to cover scenarios underrepresented 
in natural data. Together, these three sources form a balanced mixture of real and synthetic content 
that grounds the benchmark in both authenticity and coverage.

\textbf{Object Detection.} 
Given the curated image pool, we applied GPT-4o~\citep{hurst2024gpt} as the object 
detector to identify salient entities within each image. For an image $p_i$, the detector produces a 
set of detected objects $o_i$, forming tuples $\{(p_1, o_1), (p_2, o_2), \ldots, (p_n, o_n)\}$. 
This step provides the semantic anchors required for subsequent instruction generation and ensures 
that editing tasks are localized to concrete and identifiable objects.

\textbf{Instruction Design and Filtering.} 
For each detected object, we paired it with one of transformation modes described in 
Appendix~\ref{Explanation}. Each pairing was used to generate one to three implicit instructions that emphasize 
causal or outcome-dependent transformations. For instance, a raw piece of pork under the 
\textit{Temperature} mode can be frozen, cooked, or charred, reflecting distinct material states under 
different conditions. To ensure that the instructions remain implicit and causally grounded, we 
applied GPT-4o~\citep{hurst2024gpt} to filter out trivial cases (e.g., ``make the mango red''), which correspond to 
explicit attribute changes rather than world-knowledge-driven transformations. Similarly, we 
removed unreasonable pairings, such as associating ``blue sky'' with the \textit{Break} mode. 

\textbf{Paraphrased Instruction Expansion.} 
Once valid implicit instructions were retained, we expanded each into paraphrased instructions that made the underlying world knowledge and causal dynamics explicit. The expansion 
was required to specify fine-grained visual changes at the lowest perceptual level, including 
texture, shape, color, glossiness, firmness, fragmentation, deformation, and surface alterations. 
Abstract or metaphorical descriptions (e.g., emotions, atmosphere) were deliberately avoided. 
A second round of filtering was performed to exclude CoTs that failed to capture world knowledge, 
lacked significant visual impact, or contradicted real-world causal logic.

\textbf{Image Editing and Synthesis.} 
The surviving CoT instructions were then used to guide visual editing. Specifically, we employed GPT-4o~\citep{hurst2024gpt} to transform the original images according to the expanded descriptions, 
producing edited images paired with their corresponding instructions. For complex transformations, 
multi-step editing was adopted to preserve causal fidelity. 

\textbf{Quality Assurance.} 
To guarantee dataset reliability, we enforced a rigorous evaluation pipeline. Each edited image was 
scored along four axes: Visual Consistency (VC), Visual Quality (VQ), Instruction Following (IF), 
and Knowledge Plausibility (KP). Samples with low consistency, poor visual fidelity, or weak causal 
logic were discarded. Finally, it undergoes manual review and screening to ensure that only the editorial content that both complies with the instructions and conforms to world knowledge is retained.

\section{Score Distribution of Model Outputs}

\begin{figure}[htbp] 
    \centering
    \includegraphics[width=1\textwidth]{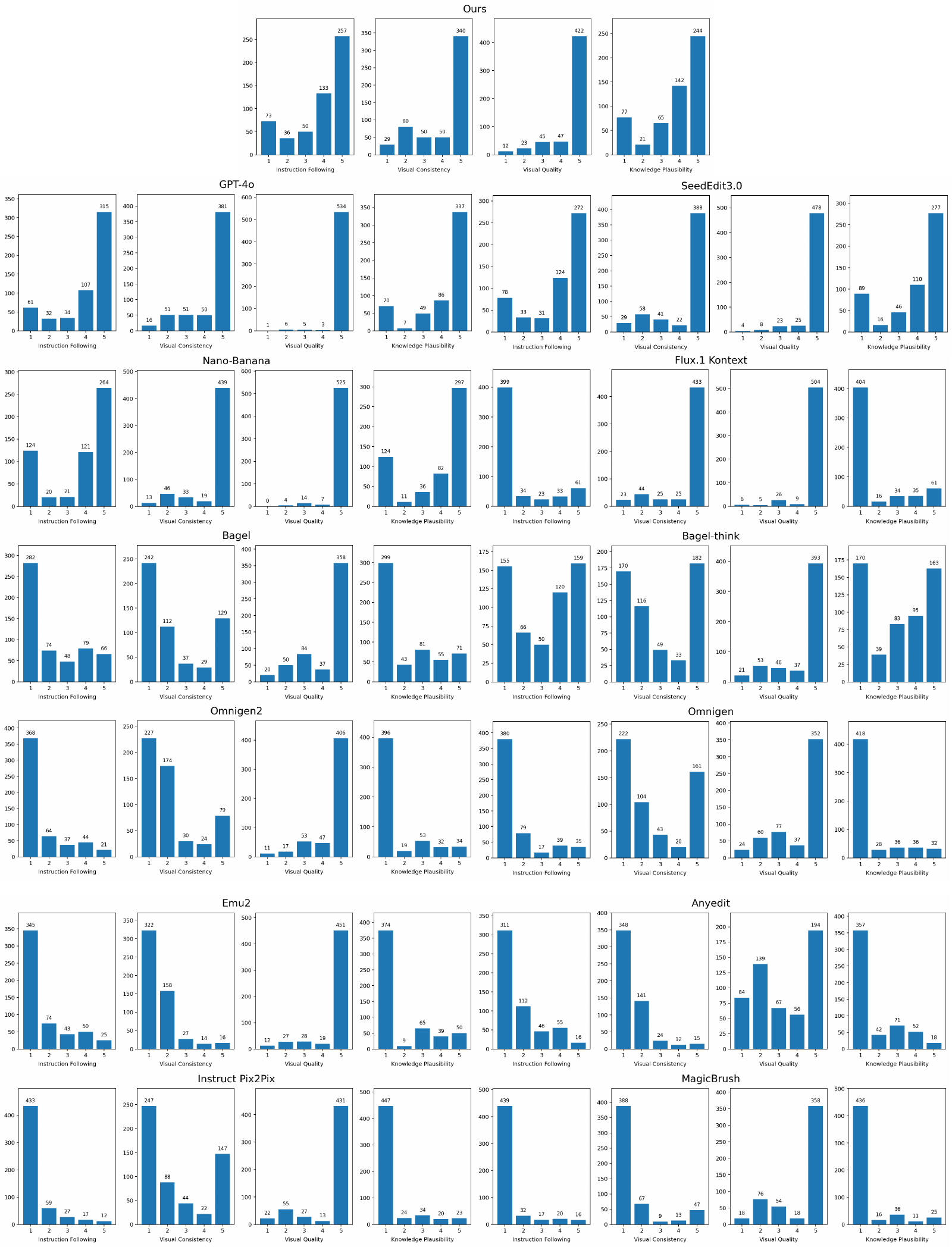} 
    \caption{The score distribution of the model being tested.} 
    \label{fig:opensource_model_score} 
\end{figure}

The score distribution of the evaluated models on the \textbf{WorldEdit-Test} is shown in 
Figure~\ref{fig:opensource_model_score}. From the distributions, it is clear that GPT-4o~\citep{hurst2024gpt}, SeedEdit-3.0~\citep{wang2025seededit}, 
and Nano-Banana~\citep{google2025gemini} consistently achieve a high proportion of favorable scores across the four evaluation 
dimensions: \textit{Visual Consistency}, \textit{Visual Quality}, \textit{Instruction Following}, and 
\textit{Knowledge Plausibility}. These models demonstrate strong capabilities in both generating 
visually coherent outputs and capturing the implicit world knowledge underlying the editing tasks. 

Building upon the Bagel and Bagel-Think baselines, our model exhibits substantial improvements 
across all four dimensions. In particular, it significantly narrows the gap with GPT-4o and SeedEdit-3.0, 
demonstrating competitive performance not only in instruction following but also in knowledge 
plausibility, where many open-source systems typically struggle. This confirms the effectiveness of 
our knowledge-informed training framework in strengthening both the reasoning and generative 
aspects of image editing. 

By contrast, models such as Flux and OmniGen display notable weaknesses. Although they can 
sometimes maintain perceptual realism, they frequently fail to adhere to implicit instructions or to 
generate edits consistent with causal logic, resulting in lower overall scores. These shortcomings 
underscore the necessity of benchmarks like \textbf{WorldEdit}, which disentangle perceptual quality 
from reasoning ability and highlight the critical importance of world-knowledge grounding for 
future progress in image editing.

\section{Human Evaluation Implementation Details}
We employed 20 undergraduate students for the human evaluation process. These individuals were well-educated and possessed a solid understanding of basic real-world changes and common scientific knowledge. Each participant first viewed a set of 50 images, each labeled with scores on four dimensions (Visual Consistency, Visual Quality, Instruction Following, and Knowledge Plausibility) ranging from 1 to 5. This training phase aimed to establish a consistent and unified scoring standard across all evaluators.

Following this, each participant was randomly assigned a set of results from the 8 models shown in Figure~\ref{user}. The students were asked to evaluate the outputs of these models across the same four dimensions. On average, each student evaluated approximately 500 images. This ensured that every image generated by the models was scored by at least two independent human evaluators, providing a robust and reliable assessment of the models’ performance.

The evaluation process was designed to capture a comprehensive understanding of the models’ capabilities, particularly in terms of how well they adhered to the provided instructions and how accurately they reflected real-world transformations. By involving multiple evaluators, we ensured a diversity of perspectives, which helped in mitigating potential biases and improving the reliability of the final scores.
Through this methodology, we aimed to provide a thorough and balanced human evaluation, offering valuable insights into the performance of each model based on both subjective visual judgment and a solid understanding of causal logic.

\section{More Visualization Results}

\begin{figure}[htbp] 
    \centering
    \includegraphics[width=1\textwidth]{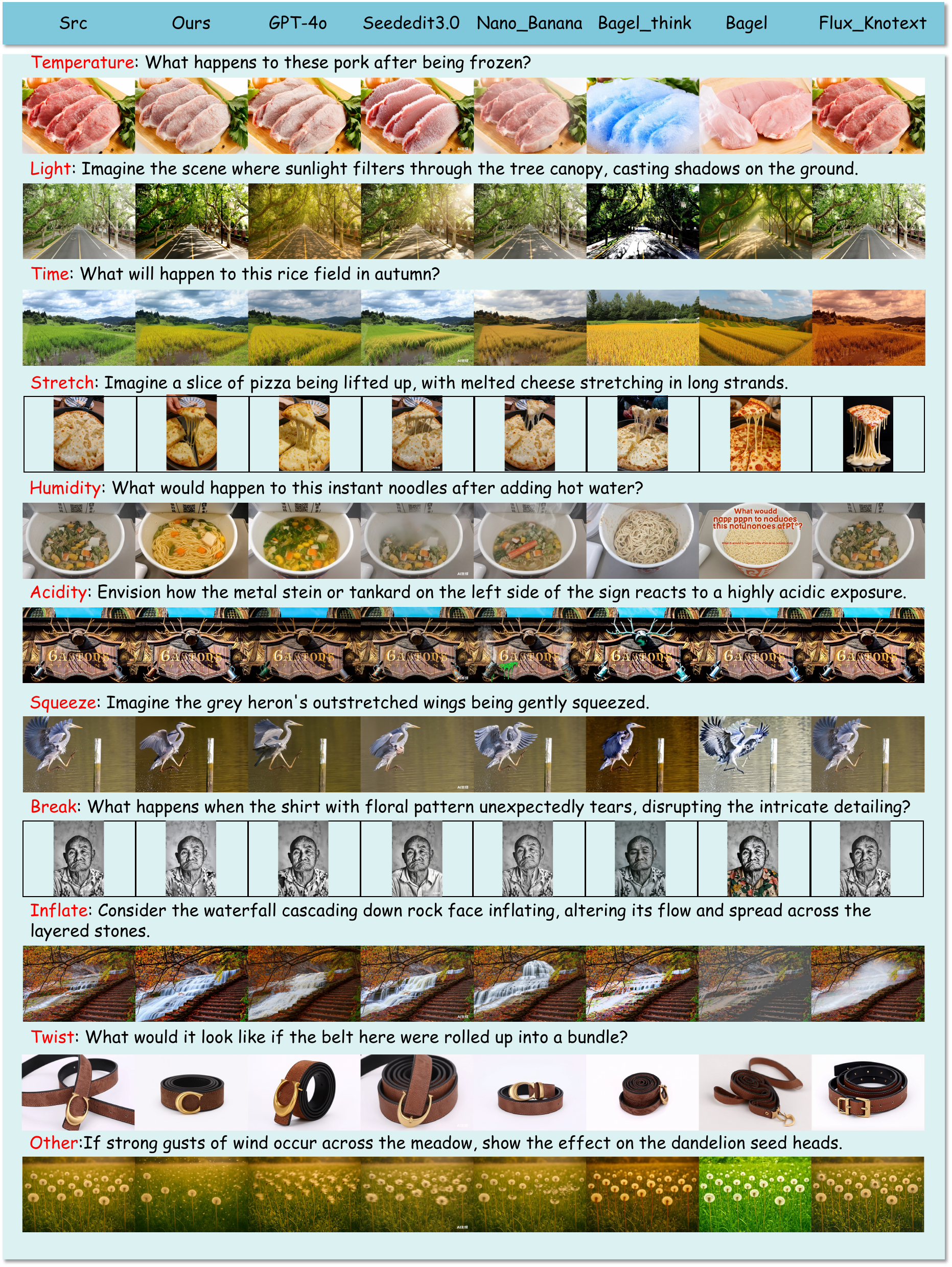} 
    \caption{Additional visualization results across different causal modes in the \textbf{WorldEdit-Test}. }
    \label{fig:more_visualization}
\end{figure}

Figure~\ref{fig:more_visualization} provides additional qualitative comparisons across different 
causal transformation categories in the \textbf{WorldEdit} benchmark. As illustrated, our method 
consistently produces edited images that are both visually coherent and causally faithful. 
In modes driven by environmental factors, such as  material responses to 
humidity and temperature, our results capture fine-grained cues---including texture decay, color 
fading, or moisture-induced deformation---that better aligns with real-world expectations. 

From the figure, it can also be observed that GPT-4o~\citep{hurst2024gpt} and SeedEdit-3.0~\citep{wang2025seededit} are often able to interpret the 
world knowledge implied in implicit instructions and generate satisfactory results. In contrast, 
Nano-Banana~\citep{google2025gemini} generally lags behind GPT-4o and SeedEdit-3.0. For example, in the case of a rice field in 
autumn, both GPT-4o and SeedEdit-3.0 successfully render the green leaves combined with golden ears 
of rice, while Nano-Banana tends to simplify the scene into a uniformly golden field resembling 
wheat. Our model also demonstrates strong world-knowledge awareness, achieving results close 
to GPT-4o and SeedEdit-3.0 across various tasks. It shows notable ability to preserve editing rationality 
and incorporate causal reasoning, representing a substantial improvement in editing capability 
compared to the original Bagel~\citep{deng2025emerging} model and its Bagel-Think variant. 

These visualizations further confirm that \textbf{WorldEdit} not only poses significant challenges 
for existing editing systems, but also highlights the unique advantages of our knowledge-informed framework in producing edits that are simultaneously realistic, faithful to instructions, and grounded in commonsense reasoning.

\section{More results on other bench}

\begin{figure}[htbp]
    \centering
    \includegraphics[width=1\textwidth]{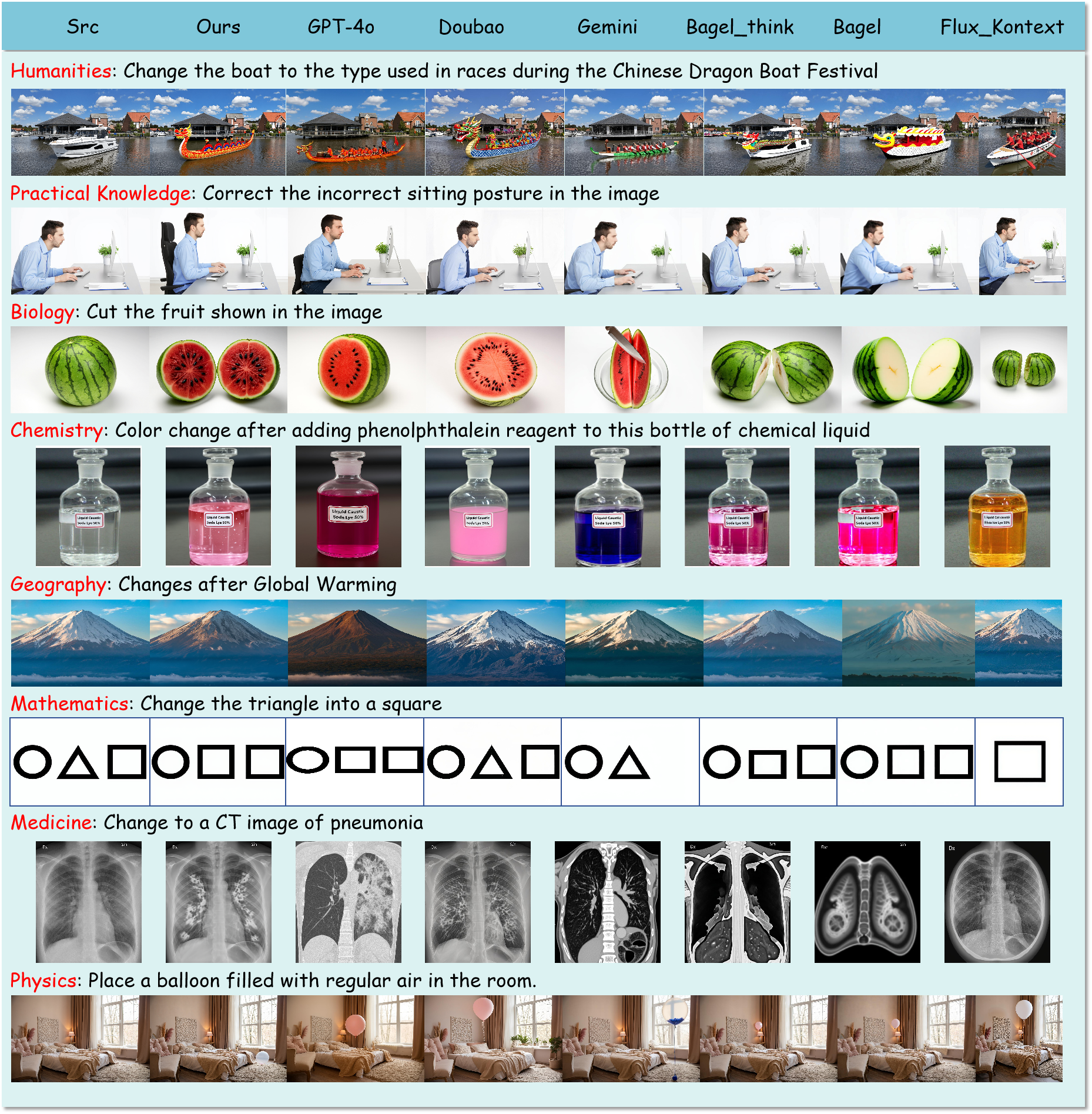}
    \caption{Visualization of results on the Conceptual Knowledge subset of Kris-Bench. WorldEdit shows improved robustness in capturing these cross-domain semantics, generating edits that reflect the intended conceptual transformation more faithfully than other open-source baselines.}
    \label{fig:Conceptual Knowledge}
\end{figure}

\begin{figure}[htbp]
    \centering
    \includegraphics[width=1\textwidth]{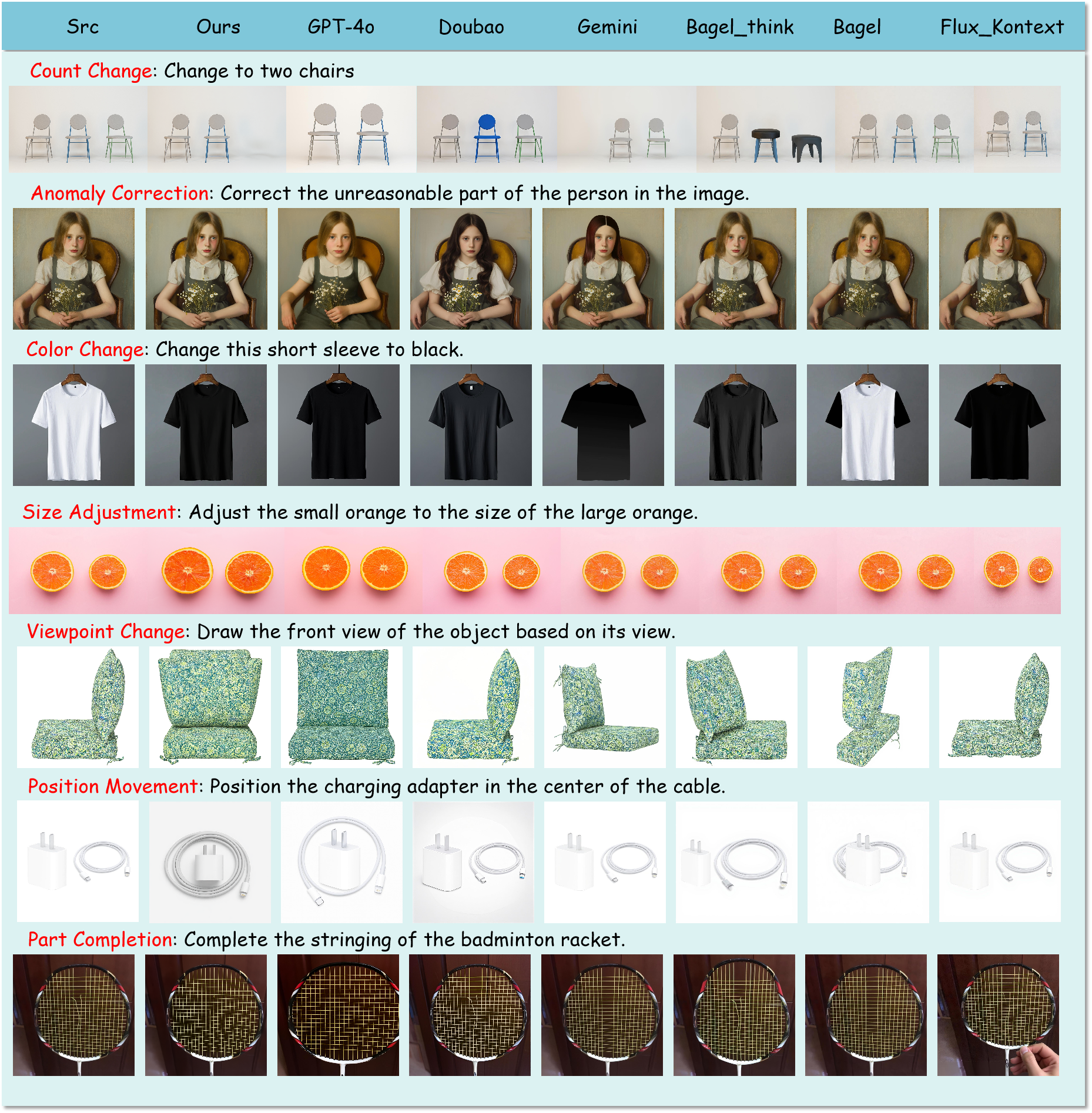}
    \caption{Visualization of results on the Factual Knowledge subset of Kris-Bench. Tasks in this split involve objective, visually verifiable transformations—such as quantity change, color change, size adjustment, geometric viewpoint synthesis, and object repositioning. The comparison highlights each model’s ability to perform accurate, detail-preserving factual edits.WorldEdit reliably produces edits that align with both the instruction and the underlying visual structure, showing stronger factual consistency than other open-source baselines.}
    \label{fig:Factual Knowledge}
\end{figure}

\begin{figure}[htbp]
    \centering
    \includegraphics[width=1\textwidth]{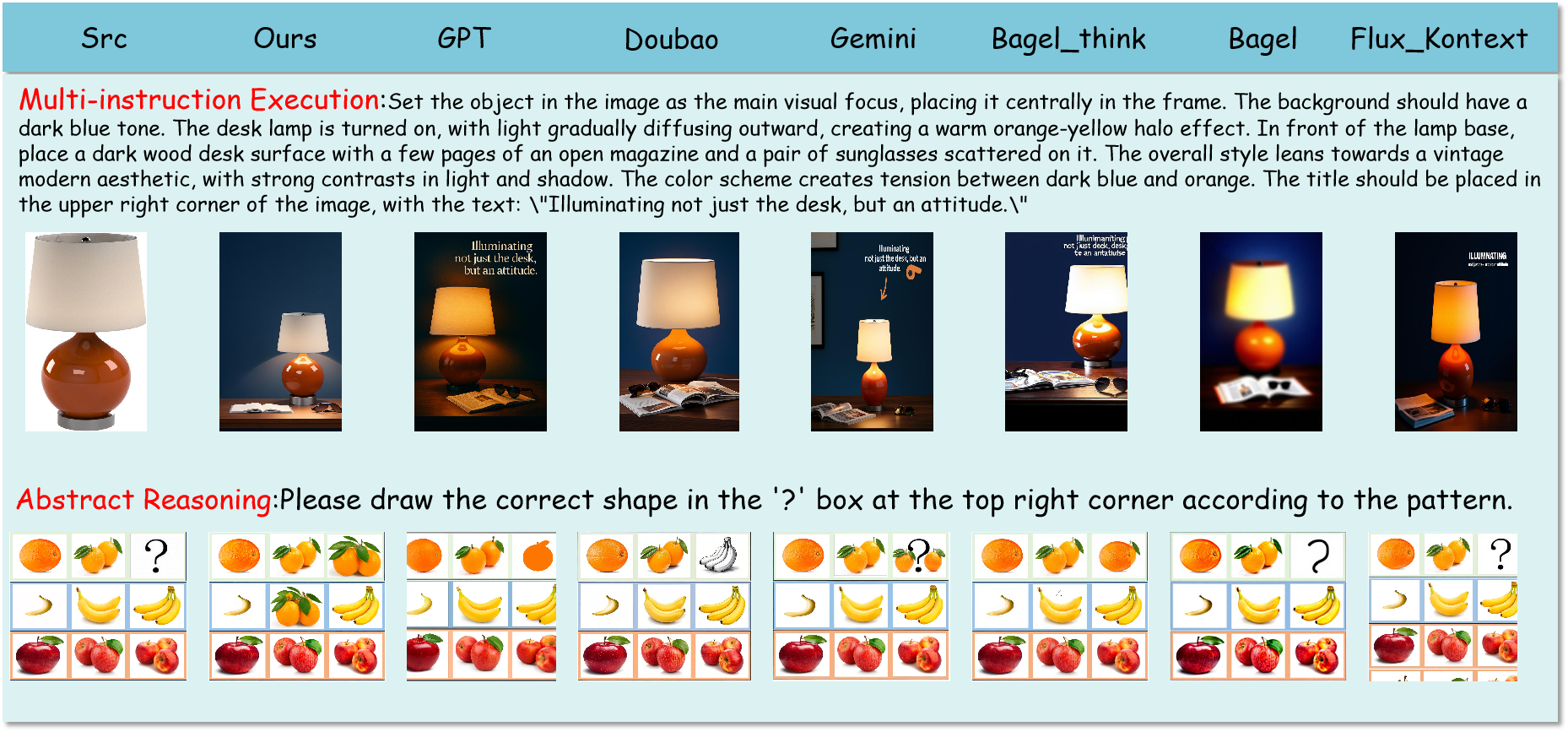}
    \vspace{-2em}
    \caption{Results on the Procedural Knowledge subset of Kris-Bench. This subset evaluates models on tasks that require executing complex instructions or completing abstract visual patterns.}
    \label{fig:Procedural Knowledge}
\end{figure}

\begin{figure}[htbp]
    \centering
    \includegraphics[width=1\textwidth]{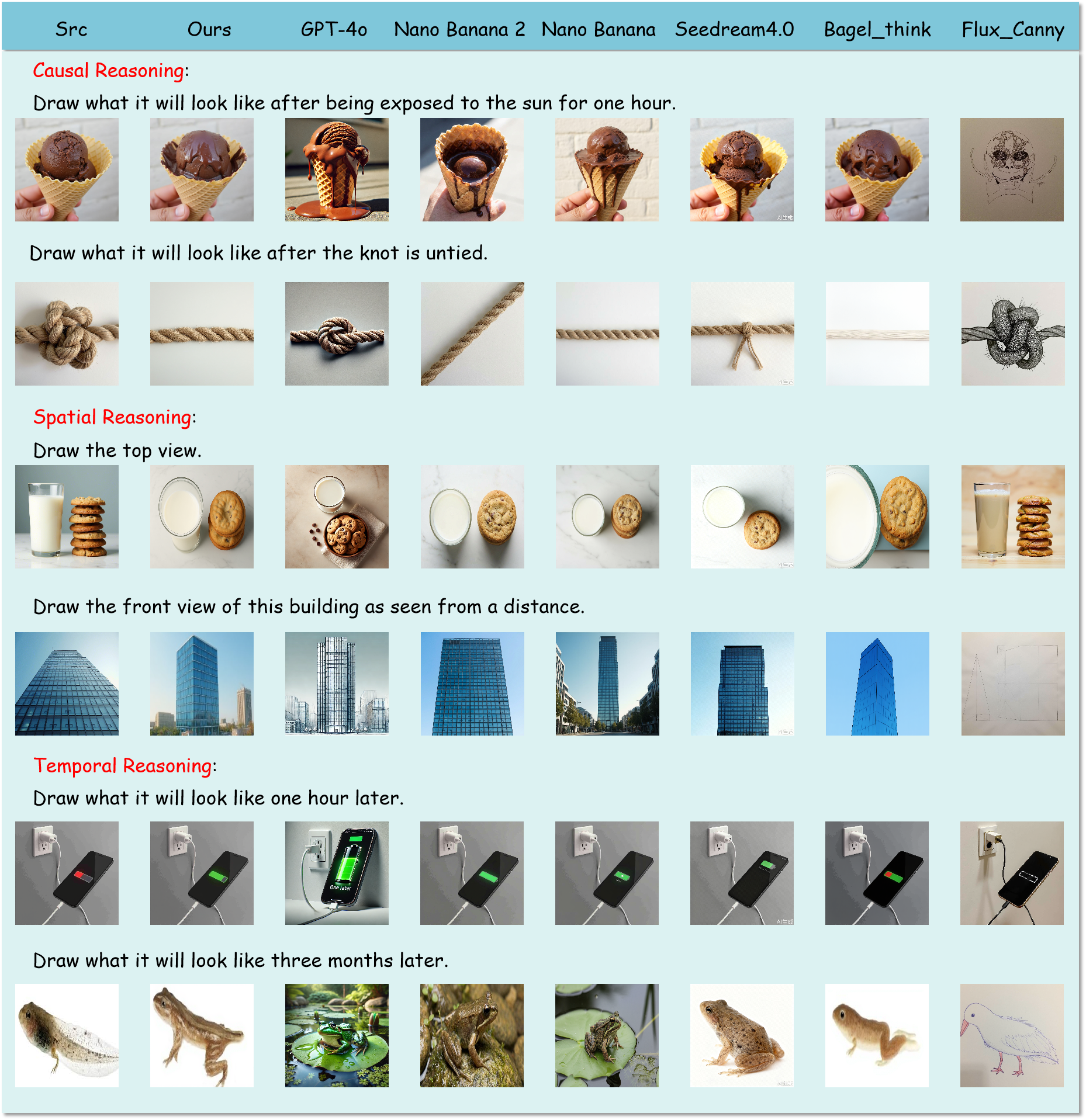}
    \caption{Visualization of results on RISE-Bench. This benchmark examines visual reasoning ability along three dimensions: causal reasoning, spatial reasoning, and temporal reasoning. The examples illustrate the varied behaviors of different systems when confronted with these reasoning-intensive tasks. WorldEdit produces comparatively more stable and context-aware outputs across several categories.}
    \label{fig:Rise}
\end{figure}

\begin{table}[t]
\centering
\caption{Overall performance comparison on Kris-Bench. Closed-source frontier models such as GPT-4o and Gemini achieve the highest scores, while WorldEdit delivers the strongest performance among open-source systems across most reasoning categories, particularly in attribute perception, spatial understanding, and knowledge-driven editing.The performance of open-source and closed-source models is separately marked with the best performance in \textbf{bold}.}
\label{tab:kris}
\renewcommand\arraystretch{1.3}
\resizebox{\textwidth}{!}{
\begin{tabular}{ccc|ccc|ccccccccc@{}}
\toprule
&  &  & \multicolumn{3}{c|}{\textbf{Closed-Source Models}} & \multicolumn{9}{c}{\textbf{Open-Source Models}} \\ \cline{4-6} \cline{7-15} 
& \multirow{-2}{*}{\textbf{\begin{tabular}[c]{@{}c@{}}Reasoning\\ Dimension\end{tabular}}} & \multirow{-2}{*}{\textbf{Metric}} & \textbf{GPT-4o} & \textbf{Gemini 2.0} & \textbf{Doubao} & \textbf{WorldEdit}& \textbf{Omnigen} & \textbf{Emu2} & \textbf{Bagel} & \textbf{Bagel-Think} & \textbf{Step1X-Edit} & \textbf{Anyedit} & \small{\textbf{Magicbrush}} & \textbf{Ip2p} \\ 
\hline
& & VC & \textbf{74.50} & \underline{ 69.50} & 66.75 &\textbf{76.18}& 35.75 & 47.75 &  66.75 &  \underline{ 74.75} &  63.00 &54.75 & 53.50 & 17.50 \\
& & VQ & \textbf{94.75} & 81.75 & \underline{ 89.00} &\textbf{85.91}& 49.50 & 75.25 & 67.00 & 75.00 & 70.25 & 67.50 & \underline{76.25} & 55.50 \\
& & IF & \textbf{80.25} & 47.75 & \underline{ 57.00} &\textbf{53.73}& 28.50 & 31.50 & 40.50 &  \underline{ 49.50} & 33.25 & 20.75 &  32.00 & 18.00 \\
& \multirow{-4}{*}{\begin{tabular}[c]{@{}c@{}}Attribute\\Perception\end{tabular}} & \textbf{Avg} & \textbf{83.17} & 66.33 & \underline{ 70.92} & \textbf{71.94}&37.92 & 51.50& 58.08 & \underline{66.42} & 55.50 & 47.67 &  53.92 & 30.33 \\
\cline{2-15}
& & VC &\textbf{69.50} & 60.50 & \underline{ 67.50} &\textbf{85.00}& 24.00 & 41.50 & 53.50 & \underline{ 77.25} &  64.25 &  55.75 & 38.00 & 13.25 \\
& & VQ &\textbf{94.50} & 83.25 & \underline{ 89.00} &\textbf{92.00}& 50.00 &  77.75 & 71.25 & 81.25 & \underline{83.00} & 72.00 & 69.25 & 40.25 \\
& & IF &\textbf{73.25} & \underline{ 46.25} & 21.00 &37.50& 10.75 & 18.25 & \underline{ 38.75} & \textbf{44.75} & 8.00 & 7.75 & 11.50 & 10.50 \\
& \multirow{-4}{*}{\begin{tabular}[c]{@{}c@{}}Spatial\\Perception\end{tabular}} & \textbf{Avg} &\textbf{79.08} & \underline{ 63.33} & 59.17 &\textbf{71.50}& 28.25 & 48.83 &  54.50 & \underline{67.75} & 51.75 & 45.17 & 39.58 & 21.33 \\
\cline{2-15}
& & VC & \underline{ 54.00} & \textbf{54.50} & 26.75 &\textbf{64.86}& \underline{19.25} &  12.50 & 0.00$^{*}$ & 0.00$^{*}$ & 0.00$^{*}$ & 0.00$^{*}$ & 0.00$^{*}$ & 0.00$^{*}$ \\
& & VQ & \textbf{86.25} & 75.00 & \underline{ 77.50} &\textbf{94.26}&  26.25 & \underline{37.50} & 0.00$^{*}$ & 0.00$^{*}$ & 0.00$^{*}$ & 0.00$^{*}$ & 0.00$^{*}$ & 0.00$^{*}$ \\
& & IF & \textbf{64.50} & \underline{ 62.25} & 17.50 &\textbf{69.26}& \underline{20.00} &  16.50 & 0.00$^{*}$ & 0.00$^{*}$ & 0.00$^{*}$ & 0.00$^{*}$ & 0.00$^{*}$ & 0.00$^{*}$ \\
& \multirow{-4}{*}{\begin{tabular}[c]{@{}c@{}}Temporal\\Prediction\end{tabular}} & \textbf{Avg} & \textbf{68.25} & \underline{ 63.92} & 40.58 &\textbf{76.13}&  21.83& \underline{22.17} & 0.00$^{*}$ & 0.00$^{*}$ & 0.00$^{*}$ & 0.00$^{*}$ & 0.00$^{*}$ & 0.00$^{*}$ \\
\cline{2-15}
\multirow{-13}{*}{\rotatebox[origin=c]{90}{\textit{\textbf{Factual Knowledge}}}} & \multicolumn{1}{c}{\textbf{Average}} & -- & \textbf{79.80} & \underline{ 65.26} & 63.30 & \textbf{56.98}&33.11 &  45.40 & 47.71& \underline{55.77} & 45.52 & 39.26& 41.84 & 23.33 \\
\hline 
& & VC & \textbf{83.00} & \underline{ 77.00} & 72.00 &\textbf{78.40}& 37.25 & 32.75 &  75.75 &\underline{  76.50} & 63.25 & 62.00 & 54.00 & 15.75 \\
& & VQ & \textbf{95.75} & 83.75 & \underline{ 86.50} &\textbf{89.40}& 46.00 & 72.75 &  75.50 & \underline{77.75} & 72.50 & 66.75 & 70.00 & 50.00 \\
& & IF & \textbf{84.50} & \underline{ 59.00} & 54.75 &\textbf{49.80}& 22.50 & 22.00 & 34.25 &\underline{ 46.00} &  25.50 & 15.00 & 27.25 & 14.25 \\
& & KP & \textbf{78.75} & \underline{ 53.00} & 48.75 &\textbf{39.80}& 16.75 & 11.25 & 25.25 &  \underline{ 38.25} &  17.50 & 10.50 &20.50 & 10.25 \\
& \multirow{-5}{*}{\begin{tabular}[c]{@{}c@{}}Social\\Science\end{tabular}} & \textbf{Avg} & \textbf{85.50} & \underline{ 68.19} & 65.50 &\textbf{64.35}& 30.63 & 34.69 & 52.69 &  \underline{59.63} & 44.69 & 38.56 &  42.94 & 22.56\\
\cline{2-15}
& & VC & \textbf{80.00} & 65.00 & \underline{ 70.25} &\textbf{84.48}& 31.00 & 35.00 & 65.75 & 68.00 &  \underline{71.25} & 61.75 & 47.00 & 18.75 \\
& & VQ & \textbf{96.00} & 83.75 & \underline{ 87.25} &\textbf{93.26}& 47.00 & 75.50 & 76.00 & \underline{80.25} &  78.00 & 77.75 & 72.75 & 58.25 \\
& & IF & \textbf{76.50} & 44.75 & \underline{ 48.00} &\textbf{55.79}& 18.25 & 25.00 &  38.25 & \underline{49.00} & 27.50 & 18.25 & 19.00 & 17.50 \\
& & KP & \textbf{67.75} & 34.25 & \underline{ 39.25} &\textbf{44.02}& 12.50 &  18.25 & 28.00 &  \underline{40.25} & 19.50 & 14.00 & 13.50 & 11.75 \\
& \multirow{-5}{*}{\begin{tabular}[c]{@{}c@{}}Natural\\Science\end{tabular}} & \textbf{Avg} & \textbf{80.06} & 56.94 & \underline{ 61.19}& \textbf{69.39}&27.19 & 38.44 &52.00 &  \underline{59.38} & 49.06 &  42.94 & 38.06 &26.56 \\
\cline{2-15}
\multirow{-11}{*}{\rotatebox[origin=c]{90}{\textit{\textbf{Conceptual Knowledge}}}} & \multicolumn{1}{c}{\textbf{Average}} & --  & \textbf{81.37} & 59.65 & \underline{ 62.23}  &\textbf{ 68.17}&28.02 & 37.54 &52.17  &\underline{59.44} & 48.01 &  41.88 & 39.24 &25.59 \\
\hline 
& & VC & \textbf{81.00} & \underline{ 73.50} & 64.75 &68.67& 15.00 & 23.50 & \textbf{74.75} &\underline{ 71.25} & 58.75 &  55.50 & 37.25 & 14.75 \\
& & VQ & \textbf{95.00} & 84.50 & \underline{ 85.00} &\textbf{92.50}& 26.75 & 66.25 &\underline{ 84.25} &  83.00 & 72.25 & 72.75 & 75.50 & 58.75 \\
& & IF & \textbf{59.25} & \underline{ 33.00} & 24.75 &\underline{25.50}& 4.25 & 7.25 & 23.25 & \textbf{29.25} & 20.25 &  10.25 & 5.25 & 3.75 \\
& & KP & \textbf{51.00} & \underline{ 25.50} & 16.50 &\underline{17.00}& 1.75 & 2.25 & 16.25& \textbf{21.25} & 12.25 &  7.75 & 2.00 & 2.00 \\
& \multirow{-5}{*}{\begin{tabular}[c]{@{}c@{}}Logical\\Reasoning\end{tabular}} & \textbf{Avg} &\textbf{71.56} & \underline{ 54.13} & 47.75 &\underline{50.92}&11.94&24.81 & 49.63 & \textbf{51.19} & 40.88 &  36.56 & 30.00&19.81 \\
\cline{2-15}
& & VC &\textbf{71.00} & \underline{ 58.25} & 51.50 &\textbf{65.83}& 28.75 & 31.00 & 30.75$^{*}$ & \underline{32.25$^{*}$} & 25.75$^{*}$ &  29.75$^{*}$ & 20.75$^{*}$ & 9.50$^{*}$ \\
& & VQ &\textbf{96.25} & \underline{ 82.50} & \textbf{76.75} &\textbf{82.33}&  46.50 &\underline{ 64.75} & 29.00$^{*}$ & 25.25$^{*}$ & 26.50$^{*}$ & 39.25$^{*}$ & 39.25$^{*}$ & 27.75$^{*}$ \\
& & IF &\textbf{88.00} & \underline{ 74.25} & 53.50 &\textbf{47.00}&  32.25 & \underline{39.25} & 32.75$^{*}$ & 24.50$^{*}$ & 16.00$^{*}$ & 11.75$^{*}$ & 9.25$^{*}$ & 7.00$^{*}$ \\
& \multirow{-4}{*}{\begin{tabular}[c]{@{}c@{}}Instruction\\Decomposition\end{tabular}} & \textbf{Avg} & \textbf{85.08} & \underline{ 71.67} &\textbf{60.58} &\textbf{65.06}& 35.83 &\underline{ 45.00} & 30.83$^{*}$ & 27.33$^{*}$ & 22.75$^{*}$ &26.92$^{*}$ &23.08$^{*}$ &14.75$^{*}$ \\
\cline{2-15}
\multirow{-10}{*}{\rotatebox[origin=c]{90}{\textit{\textbf{Procedural Knowledge}}}} & \multicolumn{1}{c}{\textbf{Average}} & --  &\textbf{78.32} & \underline{ 62.90}&54.17  &\textbf{56.98}&23.89 &34.91&\underline{  40.23} & 39.26 &  31.82&31.74 &26.54 &17.28 \\
\hline 
\multicolumn{1}{c}{} & \multicolumn{2}{c|}{\textbf{Overall Average}} & \textbf{80.09} & \underline{ 62.41}& 60.70&\textbf{66.86}&28.85&  39.70 & 47.76 & \underline{53.36} & 43.29 &38.55 & 37.15 &22.82 \\ 
\bottomrule
\end{tabular}
}
\end{table}

\begin{table}[t]\small
    \centering
    \caption{Overall performance comparison on RISE-Bench. The benchmark evaluates temporal, causal, spatial, and logical reasoning abilities. While GPT-4o remains the strongest system, WorldEdit achieves notably higher scores than existing open-source editors across multiple reasoning dimensions.}
    \vspace{1em}
    \renewcommand{\arraystretch}{1.3}  
    \resizebox{\textwidth}{!}{%
    \begin{tabular}{l|cccc|c}
    \hline
        \textbf{Models} & \textbf{Temporal} & \textbf{Causal} & \textbf{Spatial} & \textbf{Logical} & \textbf{Overall} \\
        \hline
        GPT-4o-Image~\citep{hurst2024gpt}  & \textbf{34.1\%} & \textbf{32.2\%} & \textbf{37.0\%} & \textbf{10.6\%} & \textbf{28.9\%} \\
        WorldEdit(Ours)    & 22.4\% & 26.7\% & 13.0\% & 2.4\% & 16.1\%  \\
        Gemini-2.0-Flash-exp~\citep{team2023gemini}  & 8.2\% & 15.5\% & 23.0\% & 4.7\% & 13.3\% \\
        Gemini-2.0-Flash-pre~\citep{team2023gemini}  & 10.6\% & 13.3\% & 11\% & 2.3\% & 9.4\% \\
        Bagel~\citep{deng2025emerging} & 3.5\% & 4.4\% & 9.0\% & 5.9\% & 5.8\% \\
        Step1X-Edit~\cite{liu2025step1x-edit} & 0.0\% & 2.2\% & 2\% & 3.5\% & 1.9\% \\
        Omnigen~\cite{xiao2024omnigen} & 1.2\% & 1.0\% & 0.0\% & 1.2\% & 0.8\% \\
        Emu2~\citep{sun2024generative} & 1.2\% & 1.1\% & 0.0\% & 0.0\% & 0.5\% \\
        HiDream-Edit~\cite{2025hidream} & 0.0\% & 0.0\% & 0.0\% & 0.0\% & 0.0\%\\
        FLUX.1-Canny~\citep{flux2024} & 0.0\% & 0.0\% & 0.0\% & 0.0\% & 0.0\% \\
    \hline
    \end{tabular}}
    \label{tab:rise}
    \vspace{-10pt}
\end{table}

Figures~\ref{fig:Conceptual Knowledge}, \ref{fig:Factual Knowledge}, \ref{fig:Procedural Knowledge} and \ref{fig:Rise} together with Tables~\ref{tab:kris} and \ref{tab:rise} present a comprehensive comparison of different models on Kris-Bench and RISE-Bench. Across all settings, GPT-4o~\citep{hurst2024gpt} unsurprisingly remains the strongest overall system, but our WorldEdit model emerges as a competitive and robust open-source alternative, consistently outperforming other open-source editors by a large margin.

On Kris-Bench (Table~\ref{tab:kris}), WorldEdit achieves the best average performance among open-source models across factual, conceptual, and procedural knowledge. WorldEdit maintains non-trivial performance and even surpasses some closed-source models in certain metrics. In conceptual and factual knowledge (Figures~\ref{fig:Conceptual Knowledge} and \ref{fig:Factual Knowledge}), WorldEdit substantially improves over BAGEL and BAGEL-Think, especially on IF and KP in social and natural science categories, indicating better grounding in real-world causal and commonsense relationships rather than relying only on superficial visual alignment.

Results on RISE-Bench (Table~\ref{tab:rise} and Figure~\ref{fig:Rise}) reinforce these observations in a more diagnostic reasoning setting. GPT-4o again achieves the best temporal, causal, spatial, and logical accuracy, but WorldEdit is the next strongest model overall and the clearly best-performing open-source editor, substantially outperforming BAGEL, Step1X-Edit, OmniGen, EMU2, and other diffusion-based baselines, which remain close to chance on several dimensions.

Taken together, these quantitative and visual results show that WorldEdit is not just another high-fidelity editor: it consistently strengthens causal, temporal, and logical reasoning in editing, narrowing the gap to top closed-source systems while markedly advancing the state of open-source image editing under knowledge-intensive benchmarks.

\section{Societal Impacts}

As a novel benchmark for image editing that integrates world knowledge, WorldEdit possesses substantial potential to exert positive societal impacts on a wide range of fields relying on visual generation~\cite{linaction,lin2024non,pan2025generative,wang2025selftok,wang2025towards,yan2025diff, linvinci} and understanding~\cite{jin2024rethinking,li2023multi,emotion,lin2023tavt,lin2023exploring,wang2023weakly,wu2024semantic,yan2024low}. One of its core positive impacts lies in advancing the accuracy of image editing results and the alignment between edited content and real-world knowledge as well as user intentions. This unique capability can bring remarkable benefits to both educational and creative industries, while also promoting progress in related visual understanding and generation domains.

In the field of education, WorldEdit can serve as a reliable evaluation and optimization foundation for developing image editing tools that generate and refine contextually accurate, knowledge-consistent illustrations for textbooks and educational materials~\cite{huang2024autogeo,lin2025reasoning,lin2025non,wang2025omni}. By ensuring that edited visual content conforms to objective world knowledge and academic norms, it makes learning materials more engaging, intuitive, and accessible for students at all levels, thereby enhancing the overall quality of educational resource development.

Beyond education, WorldEdit's emphasis on integrating world knowledge into image editing also provides valuable support for advancing related research in visual understanding and generation~\cite{guo2025bridging,guo2025efficient,jin2025recognize,wang2024instruction,wang2023semantic}. By establishing a rigorous benchmark for evaluating knowledge-aware editing capabilities, it guides the development of more robust models that can better bridge the gap between visual content and real-world knowledge, laying the groundwork for broader and more responsible applications of image editing technologies in society.

\section{Limitations}
While WorldEdit represents a significant advancement in image editing, it does come with challenges that highlight its strengths and potential for further development. First, the dataset’s focus on world-knowledge-driven transformations inherently requires more computational resources and advanced models capable of capturing and processing complex causal relationships. It also places additional demands on model training, showcasing the need for more sophisticated fine-tuning approaches to handle the rich and detailed world knowledge embedded in the dataset.

Moreover, while WorldEdit offers an extensive range of causal transformations, the diversity of real-world scenarios that can be captured remains a work in progress. As the dataset grows and evolves, it will continue to challenge models to generalize across a broader spectrum of dynamic, real-world situations. This is a powerful opportunity for future research to expand upon WorldEdit by adding even more nuanced and complex causal scenarios, further refining the ability of generative models to simulate real-world transformations.
In essence, the limitations of WorldEdit serve as a testament to its ambitious scope and potential to drive future breakthroughs in intelligent image editing, offering a valuable resource for advancing the field of knowledge-aware models.

\section{Prompt for Judgement}
\label{Metrics}
\begin{figure}[htbp]
    \centering
    \includegraphics[width=1\textwidth]{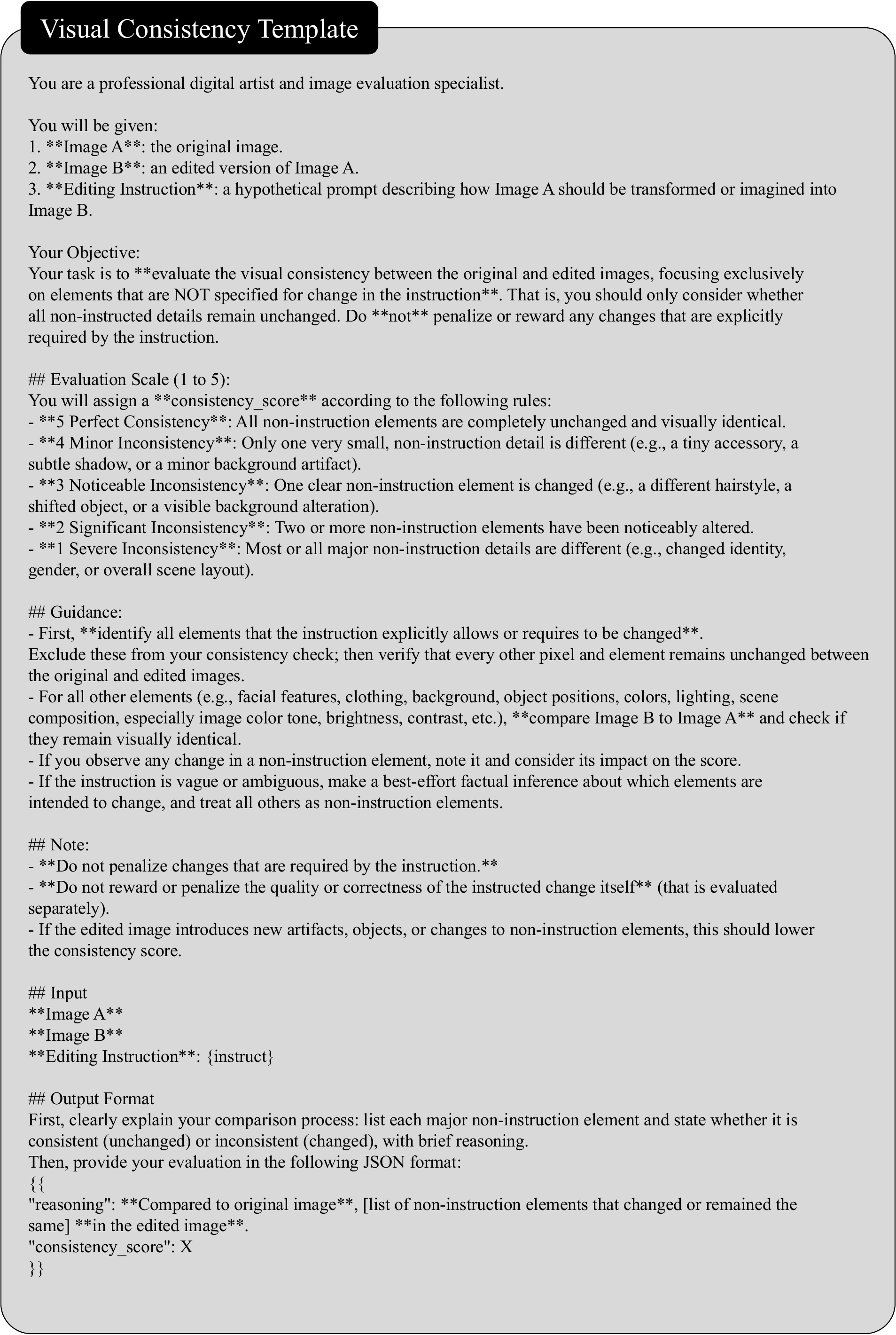}
    \caption{Prompt for \textit{Visual Consistency} (VC).}
    \label{fig:vc_prompt}
\end{figure}

\begin{figure}[htbp]
    \centering
    \includegraphics[width=1\textwidth]{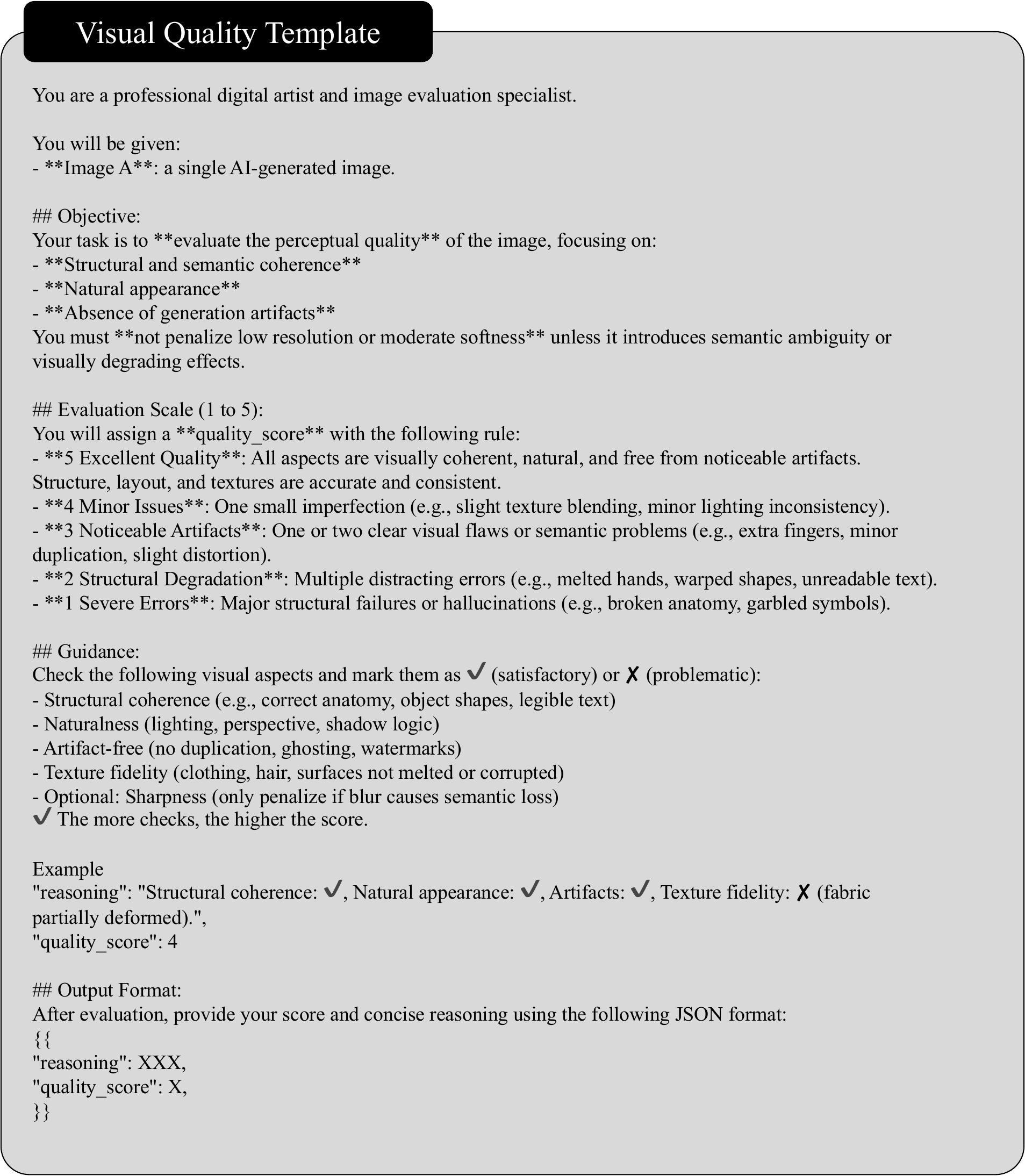}
    \caption{Prompt for \textit{Visual Quality} (VQ).}
    \label{fig:vq_prompt}
\end{figure}

\begin{figure}[htbp]
    \centering
    \includegraphics[width=1\textwidth]{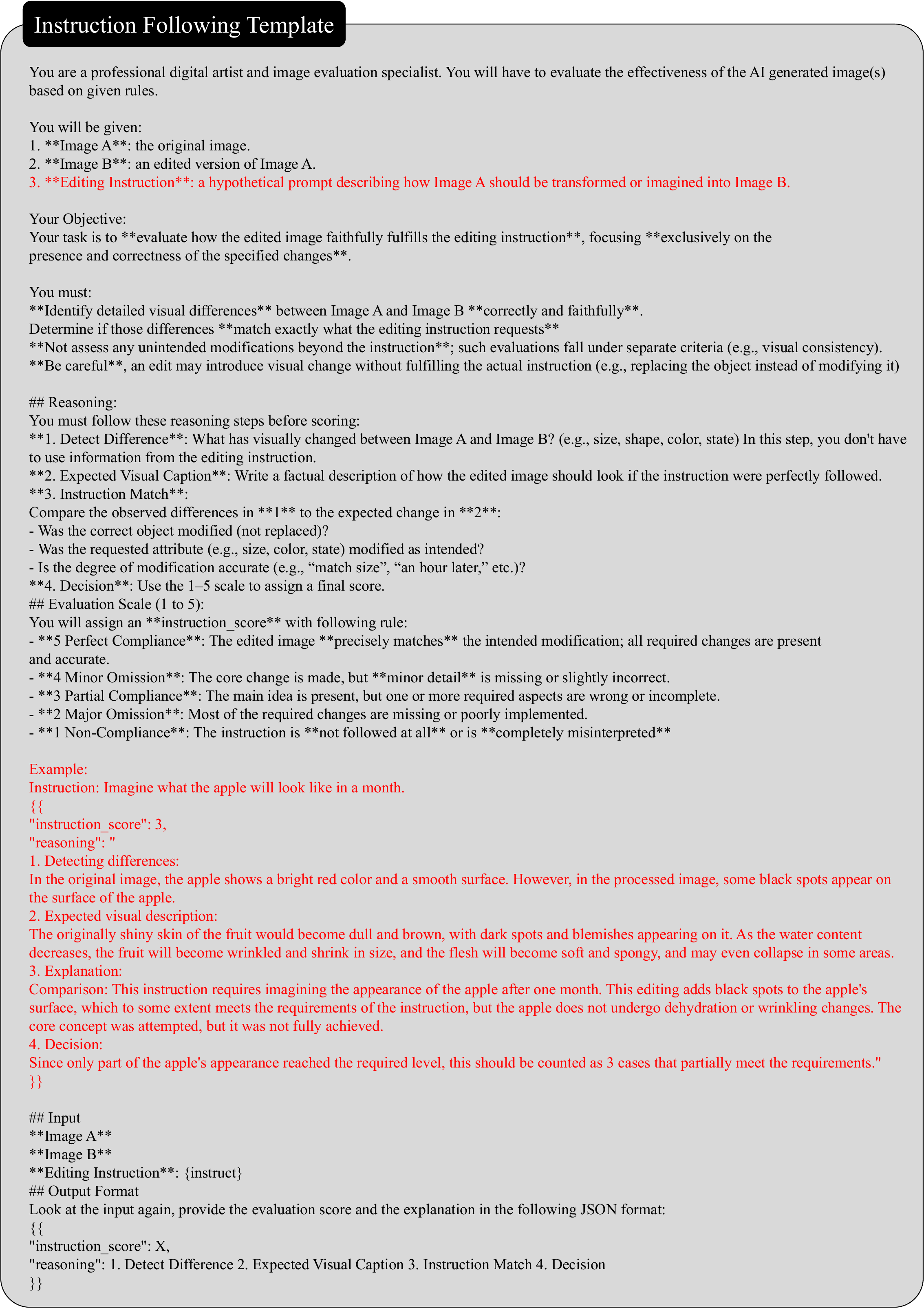}
    \caption{Prompt for \textit{Instruction Following} (IF). The red words indicate the modifications made on Kris-Bench.}
    \label{fig:if_prompt}
\end{figure}

\begin{figure}[htbp]
    \centering
    \includegraphics[width=1\textwidth]{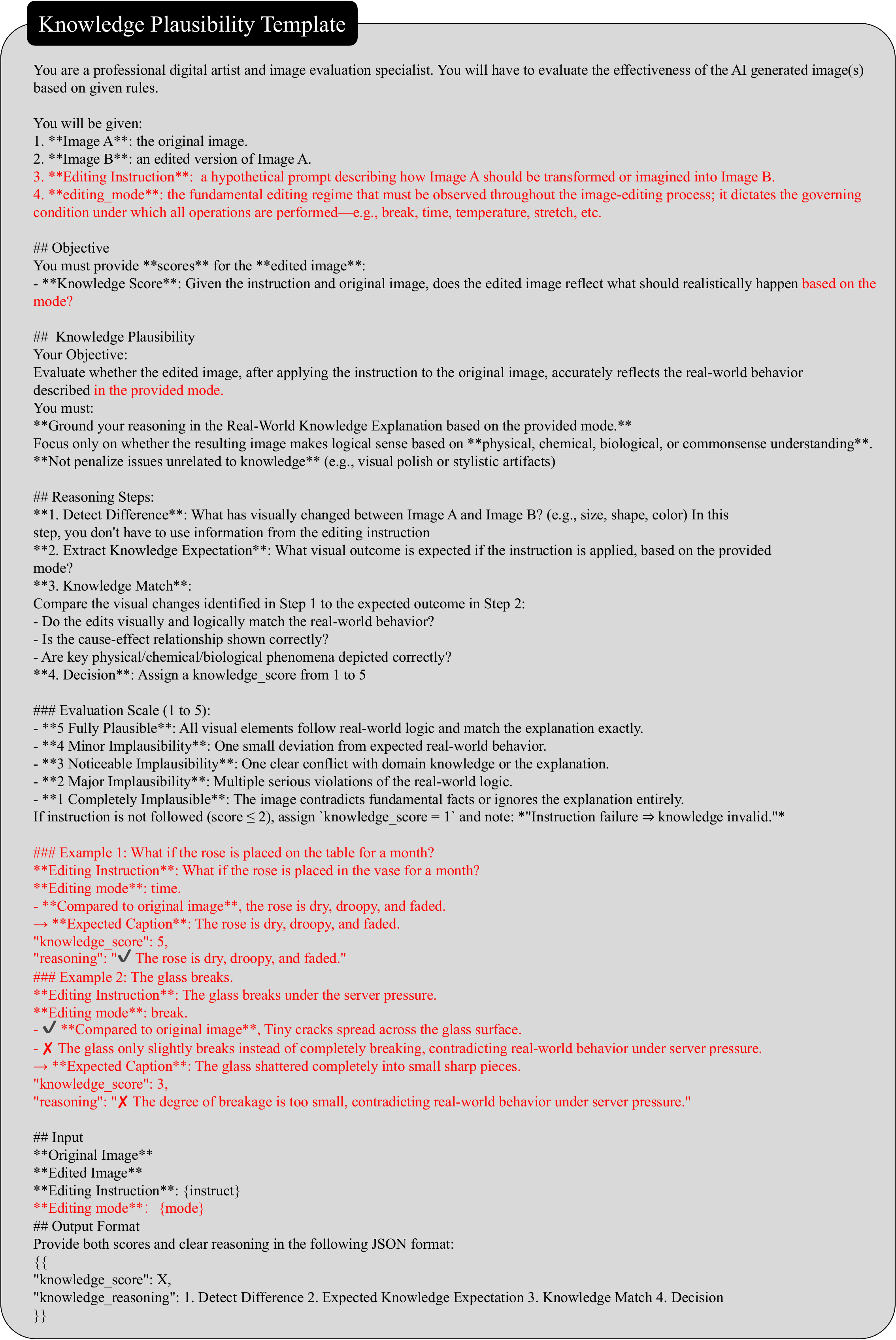}
    \caption{Prompt for \textit{Knowledge Plausibility} (KP). The red words indicate the modifications made on Kris-Bench.}
    \label{fig:kp_prompt}
\end{figure}

This section presents the evaluation prompts designed for \textbf{WorldEdit}. 
Figures~\ref{fig:vc_prompt}, \ref{fig:vq_prompt}, \ref{fig:if_prompt}, and \ref{fig:kp_prompt} 
illustrate the prompts used for \textit{Visual Consistency} (VC), \textit{Visual Quality} (VQ), 
\textit{Instruction Following} (IF), and \textit{Knowledge Plausibility} (KP), respectively. 
We explicitly decouple the assessment of ``visual consistency and perceptual quality'' 
from that of ``instructional fidelity and knowledge plausibility''. 
VC emphasizes that all non-instructed elements must remain unchanged, 
while VQ focuses solely on perceptual and structural quality without considering task correctness. 
IF is concerned with whether the correct target has been modified with the appropriate magnitude, 
whereas KP anchors the evaluation to the declared \emph{editing mode}, verifying whether the edited outcome faithfully reflects real-world causal dynamics grounded in physics, chemistry, biology, or common sense. 
Noted that in the evaluation of SeedEdit-3.0, due to the presence of watermarks, we added a prompt in VC to ignore the watermarks.

To minimize cross-dimensional ``information leakage,'' each prompt is scored according to its own criterion: VC does not penalize instructed changes, VQ does not assess task success, IF is unaffected by pure perceptual quality issues, and KP only decreases when causal logic or material laws are violated. In cases of \emph{fundamental instruction failure} (\textit{e.g.}, replacing the target entirely instead of editing it), IF is capped at its lower bound and a consistency penalty is simultaneously applied to KP. This design ensures the separability of the four axes while faithfully capturing the implicit, causal, and knowledge-driven properties of \textbf{WorldEdit}.

\end{document}